\documentclass[sn-mathphys]{sn-jnl}% Math and Physical Sciences Reference Style
\usepackage{amsmath,amssymb,amsfonts}
\usepackage{graphicx}
\usepackage{textcomp}
\usepackage{booktabs,makecell,multirow}
\usepackage{setspace}
\usepackage{threeparttable}
\usepackage{footnote}
\usepackage{makecell}
\usepackage{bbding}
\usepackage{verbatim}
\usepackage{color}
\usepackage{adjustbox}

\jyear{2021}%

\theoremstyle{thmstyleone}%
\theoremstyle{thmstyletwo}%

\theoremstyle{thmstylethree}%

\begin{document}

\title[ ]{Learning to Generalize over Subpartitions for Heterogeneity-aware Domain Adaptive Nuclei Segmentation}


\author[1]{\fnm{Jianan} \sur{Fan}}\email{jfan6480@uni.sydney.edu.au}

\author[1]{\fnm{Dongnan} \sur{Liu}}\email{dongnan.liu@sydney.edu.au}

\author[2,3]{\fnm{Hang} \sur{Chang}}\email{hchang@lbl.gov}

\author*[1]{\fnm{Weidong} \sur{Cai}}\email{tom.cai@sydney.edu.au}

\affil*[1]{\orgdiv{School of Computer Science}, \orgname{University of Sydney}, \orgaddress{\city{Sydney}, \postcode{2008}, \state{NSW}, \country{Australia}}}

\affil[2]{\orgdiv{Berkeley Biomedical Data Science Center}, \orgname{Lawrence Berkely National Laboratory}, \orgaddress{\city{Berkeley}, \postcode{94720}, \state{CA}, \country{United States}}}

\affil[3]{\orgdiv{Biological Systems and Engineering Division}, \orgname{Lawrence Berkely National Laboratory}, \orgaddress{\city{Berkeley}, \postcode{94720}, \state{CA}, \country{United States}}}


\abstract{
	%%%
	Annotation scarcity and cross-modality/stain data distribution shifts are two major obstacles hindering the application of deep learning models for nuclei analysis, which holds a broad spectrum of potential applications in digital pathology.
	Recently, unsupervised domain adaptation\;(UDA) methods have been proposed to mitigate the distributional gap between different imaging modalities for unsupervised nuclei segmentation in histopathology images.
	%However, current state-of-the-art methods neglect the severe heterogeneity within the histopathology image domain incurred by mixed cancer types and propose to align the entire domain integrally.
	%Considering that UDA methods rely on the assumption that each domain should have uni-modal data distribution, when applied in this case, they are typically prone to derive a biased alignment and consequently result in degraded performance.
	However, existing UDA methods are built upon the assumption that data distributions within each domain should be uniform.
	Based on the over-simplified supposition, they propose to align the histopathology target domain with the source domain integrally, neglecting severe intra-domain discrepancy over subpartitions incurred by mixed cancer types and sampling organs.
	%\textcolor{blue}{When the histopathology images in target domain have distinct distributions, UDA methods are inevitably prone to derive a biased alignment and compromise the efficacy of cross-domain generalization.}
	In this paper, for the first time, we propose to explicitly consider the heterogeneity within the histopathology domain and introduce open compound domain adaptation\;(OCDA) to resolve the crux.
	In specific, a two-stage disentanglement framework is proposed to acquire domain-invariant feature representations at both image and instance levels. 
	The holistic design addresses the limitations of existing OCDA approaches which struggle to capture instance-wise variations.
	%A progressive clustering and separation strategy and a global-local style consistency mechanism
	Two regularization strategies are specifically devised herein to leverage the rich subpartition-specific characteristics in histopathology images and facilitate subdomain decomposition.
	Moreover, we propose a dual-branch nucleus shape and structure preserving module to prevent nucleus over-generation and deformation in the synthesized images.
	Experimental results on both cross-modality and cross-stain scenarios over a broad range of diverse datasets demonstrate the superiority of our method compared with state-of-the-art UDA and OCDA methods.
}

\keywords{Unsupervised domain adaptation, Nuclei instance segmentation, Open compound domain adaptation, Heterogeneity}



\maketitle

\section{Introduction}\label{sec1}
%\IEEEPARstart{T}{hanks} to the progress of whole slide imaging techniques, the field of digital pathology has witnessed spectacular developments \cite{b1}. 
%In particular, it is considered as a powerful and effective tool in clinical cancer diagnostics and cancer research \cite{b2}. 
Nuclei instance segmentation, which demands both accurate localization and precise boundary delineation of each cell nucleus, plays an essential role in computer-aided digital pathology analysis\;\citep{Dey2010}. 
It captures rich characteristics of cell nuclei clusters, including their spatial distribution information and pleomorphic features, to comprehensively represent the properties of the tumor microenvrionment and is thus valuable for various clinical tasks, such as cancer identification and grading\;\citep{Dunne2001, Veta2012, baruvcic2022characterization}.

%Maybe another few lines to illustrate the difficulty of nuclei instance segmentation.
%Despite of its importance and necessity, nuclei instance segmentation is very complicated and still faces several challenges.
%First, given the surge demand of pathological examination and the tremendous amount of nuclei required to be delineate, it is impractical to follow the conventional procedure and annotate all the nuclei manually. Automated computer-aided nuclei instance segmentation methods are therefore in desperate need. 
%Besides, traditional hand-crafted features based methods suffer from a few issues in histopathology images, such as the ambiguous boundary between nuclei and tissue background, the significant inter-nuclei variations, and the clustered and overlapping nature of nuclei distribution. 
%To this end, their performance is severely deteriorated and cannot meet the standards to be applied in clinical practice \cite{b6,b7}.

%To tackle the aforementioned obstacles, deep learning-based methods are recently raised as a popular line of research and have shown remarkable success in nuclei instance segmentation \cite{b8,b9,b10}. 
Recently, deep learning-based methods have been raised as a popular line of research for nuclei instance segmentation\;\citep{Raza2019, ling2020analyzing, fujii2021x, Liu2021panoptic, mertanova2022learning}. 
Nevertheless, these methods still have non-negligible weaknesses that they heavily depend on elaborate labeled images for fully-supervised model training\;\citep{he2021cdnet, feng2021mutual}, and their performance degrades drastically under data distribution shifts (also known as domain shifts, e.g., changes in imaging modality, staining technique and cancer type between training and testing data\;\citep{Hou2019, Liu2020un}). 
%As a consequence, when a trained model is applied to a new unseen target domain, abundant annotated target domain images are still required to maintain the segmentation performance, which is unfortunately labor-consuming and often infeasible\;\citep{Torralba2011, zheng2021rectifying}.

%To alleviate the domain shift issue and maintain the label-efficiency on the target domain, unsupervised domain adaptation (UDA) method, which trains a model on the labeled source and unlabeled target domain, has recently gained a lot of traction and is regarded as a promising solution \citep{Kouw2019}. UDA methods have also demonstrated impressive efficacy in a wide range of medical image analysis tasks \citep{Cheplygina2019, Guan2021}.
A promising solution is to introduce unsupervised domain adaptation (UDA) method, which trains a model on the labeled source and unlabeled target domain\;\citep{Kouw2019}. 
It has recently gained a lot of traction and been regarded as a potential solution to alleviate the domain shift issue and maintain label-efficiency\;\citep{shen2021cdtd}. 
%UDA methods have demonstrated impressive efficacy in a wide range of medical image analysis tasks\;\citep{dong2020can, Guan2021}.
Notably, there have also been several attempts to perform domain adaptive nuclei instance segmentation\;\citep{Liu2020un, Liu2020pdam, Hsu2021}. 
They performed unsupervised nuclei segmentation in histopathology images by exploiting domain-invariant knowledge from another modality (e.g., fluorescence microscopy).
%These works showcase the promise of utilizing UDA methods to handle the domain-shift dilemma for nuclei instance segmentation where an accurate object-level annotation procedure is time-consuming and laborious.
%
%In particular for nuclei instance segmentation in histopathology images, there have been several attempts to take advantage of labeled data from another imaging modality and introduce the UDA methods to alleviate the domain shift\;\citep{Liu2020un, Liu2020pdam, Hsu2021}. 
%For example,\;\citet{Liu2020un} proposes to apply both image-level and feature-level alignment in order to tackle the UDA instance segmentation challenge from fluorescence microscopy to histopathology images. 
%Following their experiment design,\;\citet{Hsu2021} further presents a two-stage framework for UDA instance segmentation, which at first utilizes a domain separation module to achieve feature-level alignment, then employs pseudo-labeling for segmentation mask refinement. They argue that their model can adapt knowledge not only from domains which share the same object definition with the target domain, but also domains with divergent label space (e.g. natural images). 
%As for quantitative comparison, they even yield performance comparable to works based on abundant labeled images and fully-supervised learning.

\begin{figure}[!t]
	\centerline{\includegraphics[width=0.7\columnwidth]{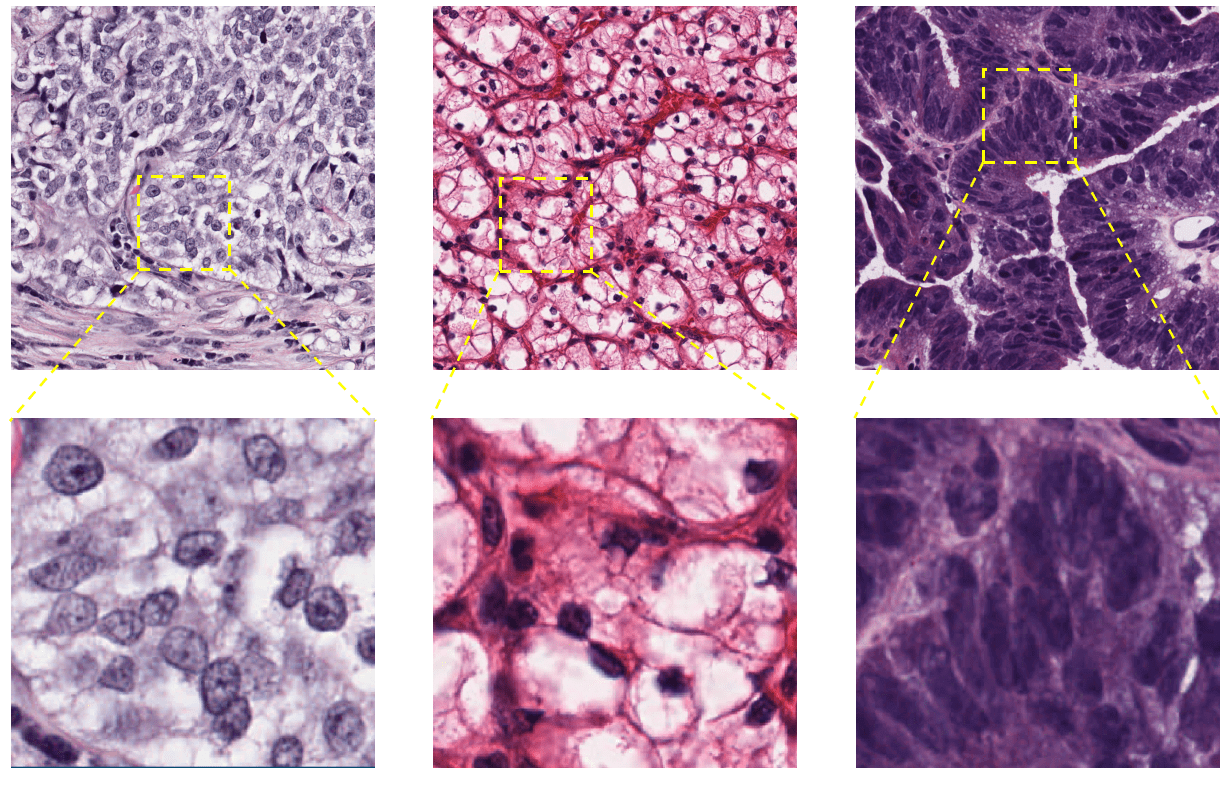}}
	\caption{Examples of histopathology images and cropped regions of different cancer types from the Kumar dataset\;\citep{Kumar2017}. From left to right: liver cancer, kidney cancer, and colon cancer.}
	\label{fig:inter-cancer}
\end{figure}

%In spite of their impressive achievements, one severe drawback of the existing approaches is that they consider all the histopathology images in the target dataset as a homogeneous domain and propose to align the entire target domain integrally with the source domain. This neglects the inevitable image appearance discrepancy and nuclei texture inconsistency between histopathology images. 
However, the existing approaches consider the target histopathology image domain as homogeneous.
They propose to align the target domain integrally with the source domain, whereas the intra-domain heterogeneity of histopathology images is neglected.
Due to inconsistent cancer types, histopathology image patches and cropped regions could exhibit diverse patterns and styles at both global image level and local instance level, as depicted in Fig.\;\ref{fig:inter-cancer}.
%As emphasized in \cite{Hou2019}, apart from imaging modality and staining technique, distinct cancer type is also a critical factor of cross-domain performance deterioration that a well-trained model can only attain inferior prediction accuracy when tested across cancer types. 
%In this regard, the current methods do not take the intra-domain heterogeneity into consideration and are thus suboptimal. Specifically, as presented in \cite{Park2020}, the conventional UDA method which is designed for uni-modal target data distribution tends to derive a biased alignment to which only target data with similar distribution to the source data can be successfully aligned.
In this case, the conventional UDA method which is designed for uniform target data distribution tends to derive a biased alignment to which only target data with similar distribution to the source data can be successfully aligned\;\cite{Park2020}.
Moreover, as these methods only regularize the model according to limited training data, they normally suffer from inferior generalization capability, especially in the realistic clinical scenario where testing images could come from divergent cancer types which do not exist in the training set.
To transcend the bottlenecks in these conventional single-source-single-target UDA approaches, it is necessary to explicitly model the heterogeneity within the histopathology image domain.

\begin{comment}
\begin{figure}[!t]
	\centerline{\includegraphics[width=0.7\columnwidth]{intra-cancer.png}}
	\caption{Examples of image patches extracted from liver cancer. It can be observed that these patches possess dissimilar image style and nuclei appearance and cannot be partitioned into one subdomain.}
	\label{fig:intra-cancer}
\end{figure}
\end{comment}

\begin{table}[!t]
	%\centering
	\caption{Comparison between OCDA and other DA settings.}
	\begin{adjustbox}{width=\columnwidth,center}
	\begin{tabular}{c|c|c|c}
		\toprule
		DA Setting & Complexity of Target Domain & Availability of Subdomain Label & Existence of Unseen Testing Subdomains\\ \midrule
		UDA & Uni-modal & --- & --- \\
		Multi-target DA & Multi-modal & \Checkmark & \XSolidBrush\\
		OCDA & Multi-modal & \XSolidBrush & \Checkmark\\
		\bottomrule
	\end{tabular}
	\end{adjustbox}
	\label{tab:setting_compare}
\end{table}

A trivial solution is to partition the whole target domain into several subdomains, following the settings of multi-target DA\;\citep{Saporta2021, Zhang2022cross}. However, such an approach has outstanding limitations that it requires domain labels to indicate the subdomain of each target sample, and it is not flexible with the complexity of target domain (i.e., the number of subdomains). 
%In histopathology image datasets, the number of cancer types is varied and acquiring subdomain-level labels still incurs annotation costs. In addition, given the conspicuous discrepancy between the cropped patches extracted from images under the same cancer type (as shown in Fig.\;\ref{fig:intra-cancer}), straightforwardly assigning subdomain labels according to cancer type is inappropriate as image patches from the same cancer type still have large distinctions. 

\begin{figure}[!t]
	\centerline{\includegraphics[width=0.9\columnwidth]{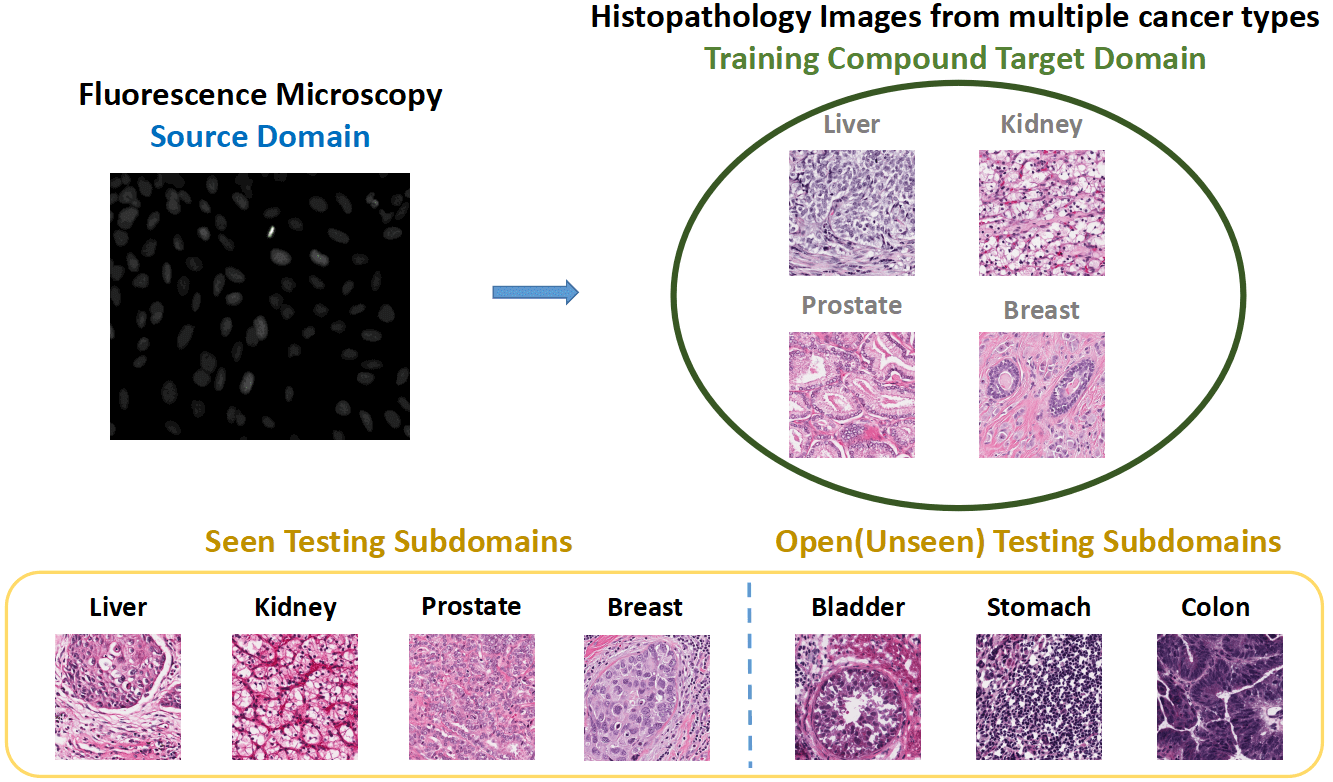}}
	\caption{Illustration of the OCDA setting in a benchmark performing domain adaptation from fluorescence microscopy to histopathology images. Note that, unlike multi-target UDA\;\citep{Saporta2021}, the cancer type of each image patch is unavailable during training.}
	\label{fig:compoundDA}
\end{figure}

In this paper, we propose a novel framework from the perspective of open compound domain adaptation (OCDA)\;\citep{Liu2020open} to address the intra-domain heterogeneity in the target histopathology dataset.
The task of this setting is to transfer knowledge from a labeled source domain to an unlabeled compound target domain, which contains multiple related yet divergent subdomains without domain labels.
In addition, the adapted model for OCDA is concurrently expected to possess better generalization capability. Therefore, the model's performance can be maintained when dealing with data from unseen subdomains at the test time, as showcased in Fig.\;\ref{fig:compoundDA}. An extensive comparison between OCDA and other UDA scenarios is illustrated in Table\;\ref{tab:setting_compare}.

OCDA is a more realistic yet relatively unexplored setting, with only a few works making an early attempt to provide a solution\;\citep{Liu2020open, Park2020, Gong2021cluster}. 
Nevertheless, they focus on down-stream tasks like image classification and semantic segmentation, where image-level semantic features are dominant. It is noted that there is an absence of OCDA framework for instance segmentation where local-level instance features are equivalently crucial and indispensible.
% Discuss why choose disentanglement-based framework.
As for technical defects, the current works mostly propose to split the compound target domain according to the style features of each sample extracted by a pre-trained model and assign unchangable domain labels at the beginning of the training stage. 
Since style feature extraction is performed via models pre-trained on other tasks, there inevitably exists noise in the encoded style representations, which causes the partition of compound target domain to be inaccurate. Then the model training at each following step would be deteriorated in consequence.
Another shortcoming of the existing methods is that they are based on an assumption that the unseen testing subdomain can be constructed as a combination of all seen training subdomains, which is actually incorrect for the histopathology image domain in regard of its complexity and countless attributes contributing to subdomain variations. 
In addition, we observe that there exists a lack of morphology-level supervision in the image synthesis framework deployed by those methods.
As a consequence, the transformed images would lose essential nucleus shape details and incur incorrespondence between images and segmentation annotations.

To this end, we propose a novel two-stage disentanglement framework to tackle nuclei instance segmentation in the OCDA setting.
It captures the domain-agnostic semantics\;(content) and the domain-specific modality/stain/cancer factors\;(style) seperately at both global image level and local instance level for mutual-complementing.
In the first image-level disentanglement stage, we present a cross-domain image translation network to transform source images to target-like ones.
In the second stage, we conduct feature disentanglement at local level to further alleviate cross-domain discrepancy in instance-level representations.
Considering the aforementioned defects of existing methods, we specifically propose four technical insights.
%Unlike existing OCDA methods, we integrate style feature extraction together with the image translation task. Specifically, we propose a progressive clustering and separation strategy to facilitate the style feature extraction during synthesis task learning.
In Stage I, firstly, we integrate the learning of style encoding together with the image translation task and propose a progressive clustering and separation strategy to facilitate style feature extraction during synthesis task learning.
%In addition, we observe that in the traditional image synthesis framework, due to the lack of morphology-level supervision, the transformed images would lose essential nucleus shape details and incur incorrespondence between images and segmentation annotations. 
Then, we seek inspiration from the recent advances of domain generalization and introduce the style randomization technique\;\citep{Jackson2019style} for data augmentation.
It strengthens the model's robustness and generalizability to maintain its performance on unseen testing subdomains.
Furthermore, we pose a dual-branch morphological regularization on top of the image translation network to minimize nucleus deformation and incorrespondence during translation.
In Stage II, we devise a global-local style consistency mechanism to stabilize the instance-level domain-invariant feature generation.

Our key contributions can be summarized as follows:
\begin{itemize}
	\item We propose a holistic two-stage disentanglement framework for cross-domain nuclei instance segmentation in the OCDA setting to explicitly address the heterogeneity of histopathology images. To the best of our knowledge, it is the first work to explicitly model the heterogeneity of histopathology images in UDA and design an OCDA framework for instance segmentation.
	% And progressive training
	\item To overcome the limitations of the existing OCDA methods, in the global image-level alignment, a progressive clustering and separation strategy is incorporated to benefit the style feature disentanglement. 
	To enhance the model's generalization capability for unseen testing subdomains, we introduce style randomization to generate fake histopathology images in arbitrary style for data augmentation.
	\item In the local instance-level alignment, we leverage the global-local style consistency to facilitate feature disentanglement and domain-invariant representation learning.
	\item We further develop a novel regularization module based on semantic masks and object boundaries to preserve shape and structural details of nucleus in image translation. 
	\item We comprehensively evaluate our approach and demonstrate its effectiveness on both cross-modality and cross-stain UDA nuclei instance segmentation. 
	It significantly outperforms the state-of-the-art conventional UDA and OCDA methods for unsupervised domain adaptive nuclei instance segmentation in histopathology images.
\end{itemize}

\section{Related work}

\subsection{Unsupervised domain adaptation}
A prominent barrier hampering the application of deep learning-based methods to healthcare is the annotated data scarcity\;\citep{nie2020adversarial, stepec2021unsupervised, han2022self, fan2023taxonomy}. 
The data collection and labeling process heavily depends on domain knowledge and requires exhaustive participation of physicians.
As a result, acquiring sufficient data with high-quality annotation could be prohibitively expensive\;\citep{cao2023autoencoder}.
Unsupervised domain adaptation (UDA) method, which aims to address the challenge by transferring domain-invariant knowledge from source domains with labeled data to unannotated target domains, has advanced rapidly and indicated its effectiveness in various applications\;\citep{dong2020can, Guan2021}.
One representative approach for solving the UDA task is through learning domain-agnostic features.
It is dedicated to mitigating the domain discrepancies by minimizing a specific metric (e.g., MMD)\;\cite{gong2014learning, yan2017mind, chen2019graph} or performing adversarial feature alignment\;\cite{tzeng2017adversarial, zhao2021madan}.
As an alternative, another line of research aims to take advantage of deep generative models\;\cite{hoffman2018cycada, li2021unsupervised, zhao2022style, benigmim2023one} or transform operations\;\cite{yang2020fda, huang2021rda, araslanov2021self, huang2021fsdr} to align different domains from the image appearance level.
An exemplary pipeline in this stream is to perform cross-domain visual mapping based on swapping of disentangled attributes \citep{Lee2020drit}.
With the same insight, following works introduce a set of ancillary constituents, such as collaborative training\;\citep{zheng2019joint, zou2020joint}, non-linear modeling\;\citep{lee2021dranet}, and identifiability constraint\;\citep{kong2022partial}, to further enhance the fidelity of disentangled representations.
%Specifically, synthetic target-like images are generated based on the source images or vice versa to compensate for the cross-domain performance deterioration.
Moreover, considering the complementary nature of feature alignment and appearance transform approaches, an integrative solution is proposed to combine them into a unified framework \cite{chen2019synergistic, Liu2020pdam, dong2020can}. By encouraging the mutual interactions and cooperations between the two perspectives of adaptations, it achieves synergistic adaptation and considerably lifts the performance.
Despite those appealing efforts, in UDA, it is assumed that both the source and target domains should strictly follow the uni-modal distribution\;\cite{zhou2022domain}.
The over-simplified paradigm cannot handle the intra-domain heterogeneity across disparate subpartitions and therefore suffer from inferior robustness in the context of multi-modal distribution\;\cite{isobe2021multi}.
%Moreover, considering the complementary nature of the feature alignment and image alignment methods, an improved solution is to integrate them into a unified framework. By encouraging the mutual interactions and cooperations between the two perspectives of adaptations, it achieves synergistic adaptation and considerably lifts the performance \cite{b54}.

%As for the feature-level UDA, despite their compelling efficacy, they propose to align the entire feature space between source and target domain, while neglecting the highly-entangled nature of those feature representations. As the key insight of feature alignment is to obtain unbiased domain-invariant representations, its performance would be seriously deteriorated if domain-specific features are not disentangled and excluded in advance \citep{Chartsias2019disentangled}. Accordingly, explicit feature disentanglement is adopted for UDA problems and intends to concretely extract domain-invariant content features and domain-specific style features. The impact and promotion of such a mechanism are comprehensively verified \citep{Meng2020mutual, Pei2021disentangle}.

\subsection{Unsupervised domain adaptation for nuclei instance segmentation}
In regard of nuclei instance segmentation in microscopy images, some pioneering works are specifically designed to handle the domain shifts in image appearance and object characteristics. 
The dominating approach is to firstly perform image translation with learning-based generative model and subsequently conduct hierarchical feature alignment\;\citep{Liu2020un}.
Auxiliary modules can be developed on top to further facilitate cross-domain generalization, such as task reweighting\;\citep{Liu2020pdam} and pseudo-labeling\;\citep{Hsu2021, li2022domain}.
However, these methods do not take the inner-discrepancy of histopathology images into consideration and instead simplify them into a homogeneous target domain. 
As a result, a biased adaptation tends to be derived and only minor subpartitions of the target domain can be reasonably aligned\;\citep{wang2019characterizing, Park2020}.
In the spirit towards a balanced and unbiased adaptation procedure, the practice followed by conventional UDA methods which conduct one-to-one alignment is inadequate\;\citep{Gong2021cluster}.
Recently, \cite{Zhang2022cross} attempted to exploit the complementarity between H\&E-stained and IHC-stained images and perform multi-target DA. 
Nevertheless, the proposed method cannot address the cancer and organ-wise heterogeneity within histopathology image domain as patch-wise subpartition labels are inaccessible\;\citep{Kumar2017}.

\subsection{Open compound domain adaptation}
Taking a step further beyond UDA with the assumption of uni-modal target data distributions, open compound domain adaptation\;(OCDA) tackles a more challenging yet practical scenario.
It models the target domain as a union of multiple subdomains and has shown appealing promises in several benchmarks of image classification\;\citep{Liu2020open} and semantic segmentation\;\citep{Park2020}.
To enhance the characterization of inner-structure with respect to the target domain, different training strategies like curriculum learning\;\citep{Liu2020open}, meta optimization\; \citep{Gong2021cluster}, and multi-teacher co-regularization\;\citep{pan2022ml} could be adopted.
%\cite{Liu2020open} presented a novel framework to investigate OCDA based on curriculum domain adaptation and memory module. 
%With a focus on OCDA for semantic segmentation, \cite{Park2020} proposed three crucial design principles and constructed a new pipeline to conduct discover, hallucinate, and adapt sequentially. 
%Comparably, \cite{Gong2021cluster} designed a four-step approach to model the compound target domain continuously and utilized meta-learning for online update. 
%\cite{pan2022ml} developed a multi-teacher learning paradigm for separate adaptation.
However, these methods possess serious shortcomings with respect to the disregard of local-level instance attributes and biased style encoding. 
In this work, we propose a holistic representation decomposition framework to bypass their limitations and pave the way for unbiased cross-domain alignment.

%\begin{comment}
\subsection{Learning-based nuclei segmentation}
In the current literature, deep learning-based methods have become prevalent in the field of nuclei instance segmentation owing to their strong feature representation capability.
These methods can be generally divided into two categories, namely proposal-free and proposal-based methods.
For the proposal-free methods, they generally follow a two-stage pipeline. 
At first, similar to the semantic segmentation task, each pixel is assigned a label to denote whether it corresponds to nuclei or tissue background. 
Then, by exploiting the spatial arrangement and morphological characteristics of the nuclei clusters, a post-processing technique is proposed to separate the overlapping nucleus entities. 
As one of the exemplar works, a deep contour-aware network\;(DCAN)\;\citep{Chen2017dcan} is formulated as a multi-task learning framework to integrate contour information with object appearance, which contributes to precise separation of the attached nuclei. 
%Additionally,\;\cite{Graham2019hover} proposed to leverage the distance measurement and constructed an auxiliary branch to perform distance regression.
%reformulate the instance segmentation task into a regression task and resorted to learn the distance-related metric. Based on the predicted distance map, post-processing techniques like watershed transformation were deployed to secure the final instance segmentation prediction.
These methods primarily rely on the global semantic characteristics yet pay less attention to the local object-level properties, and hence struggle to precisely delineate the borders between touching nuclei.
In contrast, the proposal-based methods depend on global contextual features to a less degree. 
%Specifically, an object detector is first employed to give bounding box proposals for high confidence nuclei location candidates. Subsequently, based on those proposals, regions of interest\;(ROIs) are extracted and forwarded to the instance segmentor to generate mask prediction for each object. 
They adopt the segmentation by detection procedure and constructed the segmentation branch along with the classification and box regression branches for simultaneous class, box-offset and segmentation mask predictions\;\citep{Liu2021panoptic, chen2023cpp}.
In this regard, the local instance-wise attributes can be emphasized and lead to better inter-nuclei separation. 
%However, due to the lack of global semantic information, these methods primarily focus on local structures and can only demonstrate modest segmentation performance.
%One strategy to overcome this obstacle is to bridge the global-level and local-level information together in a unified framework.
%Specifically, an end-to-end panoptic segmentation framework was designed to incorporate a semantic segmentation module into Mask R-CNN and achieved substantial improvements\;\citep{Liu2021panoptic}.}
%Extensions of this work are further performed by introducing a feature fusion mechanism to combine dual-level information and achieve substantial improvements \cite{b48,b49}.

%Considering its capability to seamlessly exploit both the global semantic and local instance features, in this work, we build our UDA model upon the panoptic segmentation architecture \cite{b47}.
%\end{comment}

\section{Methods}
\begin{figure}[!t]
	\centerline{\includegraphics[width=1.0\columnwidth]{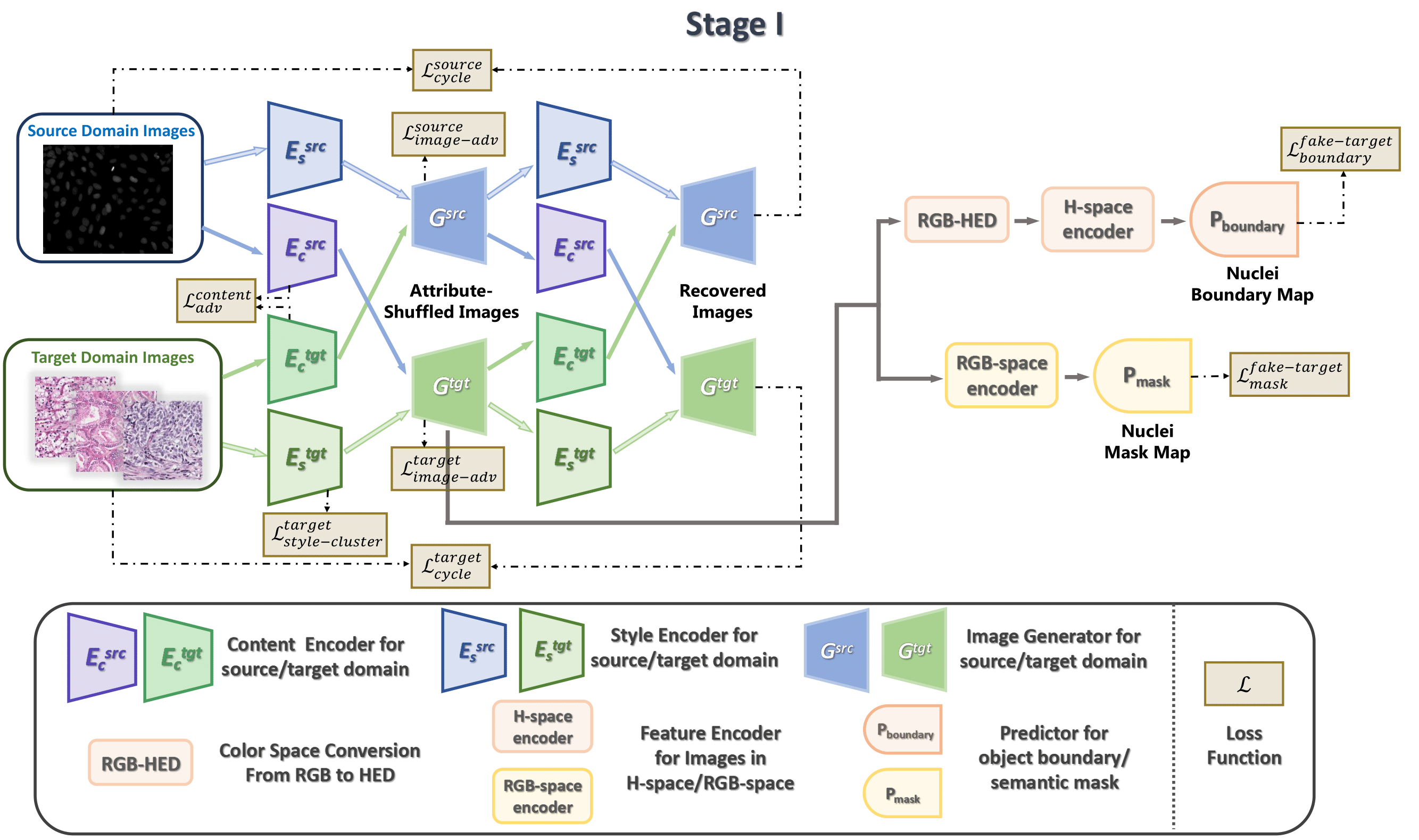}}
	\caption{Overview of Stage I for the proposed two-stage framework. The main objective of Stage I is to mitigate the significant image appearance discrepancy between different modalities and staining techniques with cross-domain image translation. A DRIT\;\citep{Lee2020drit}-like architecture is employed as backbone with several auxiliary modules to overcome its limitations.}
	\label{fig:stage1}
\end{figure}

We propose a two-stage disentanglement framework for heterogeneity-aware unsupervised domain adaptive nuclei instance segmentation from the view of OCDA setting, as demonstrated in Fig.\;\ref{fig:stage1} and Fig.\;\ref{fig:stage2}.
The model is trained sequentially such that the inference results of Stage I (i.e., synthesized target-like source images) are forwarded to Stage II as inputs. 
%The style encoder for target domain images trained in stage one is also reused in stage two with its parameters freezed.
We present the details and overall objective function of each stage in this section.

\subsection{Stage I: Cross-domain image translation with global image-level disentanglement}
To mitigate the large appearance discrepancy across images from different modalities and staining techniques, we at first propose to perform cross-domain image translation to synthesize target-like source domain images. 
Previous work\;\citep{Liu2020pdam} resorted to utilizing CycleGAN\;\citep{Zhu2017unpaired} to achieve the appearance-level adaptation. 
However, we observe that the styles of the synthesized images from CycleGAN are dominated by only one or two specific cancer styles. 
This is because CycleGAN does not explicitly model the intra-domain heterogeneity. 
%observe that since CycleGAN do not explicitly model the cancer-specific attributes and the corresponding style features, the image translation is uncontrollable and the synthesized target-like images are dominated by only one or two types of cancer in consequence.
To this end, we aim to enhance the image translation with explicit disentanglement of domain-invariant content features and domain-specific style features for more precise modeling of various cancer types. 

\subsubsection{Backbone}
Inspired by DRIT\;\citep{Lee2020drit}, we construct the framework with content encoders $E_c$, style encoders $E_s$, image generators $G$, and domain discriminators $D_{image}$ for both source and target domains, as well as a domain-invariant content discriminator $D_{content}$.
We follow the network weight sharing strategy employed in\;\citep{Lee2020drit} that the last layer of $E_c$ and the first layer of $G$ are shared across the two domains.
A disentangle-swap-reconstruct pipeline is additionally employed to regularize and guarantee the effectiveness of the feature disentanglement procedure.

To be specific, we denote the images and their corresponding annotations from source domain as $X_{src}=\left\{ (x_{src}, y_{src}) \right\}$ and the unlabeled compound histopathology target domain as $X_{tgt}=\left\{ x_{tgt}^i \right\}_{i=1}^{N_t}$, where $N_t$ indicates the number of sub-target domains which is unknown in practice.
Given an image $x_{tgt}$ from target domain, it is concurrently forwarded to the partially shared content encoder $E_c^{tgt}$ and the domain-specific style encoder $E_s^{tgt}$ to characterize its histopathological structure information $z_{c}^{tgt}$ and appearance variation $z_{s}^{tgt}$ incurred by its modality, stain, and cancer type.
Similarly, images $x_{src}$ from the labeled source domain are also encoded to extract their content and style features $\left\{z_{c}^{src}, z_{s}^{src}  \right\}$. 
Subsequently, these disentangled representations are swapped and forwarded to image generators for cross-domain image reconstruction, i.e. $X_{src}^{swap}=G^{src}(z_{c}^{tgt}, z_{s}^{src}), \  X_{tgt}^{swap}=G^{tgt}(z_{c}^{src}, z_{s}^{tgt})$.
To maintain the structure and appearance details, the disentangle-swap-reconstruct procedure is repeated on the synthesized fake images to recover the inputs as a cycle, illustrated as $X_{src}^{cycle}=G^{src}(E_c^{tgt}(X_{tgt}^{swap}), E_s^{src}(X_{src}^{swap})), \ X_{tgt}^{cycle}=G^{tgt}(E_c^{src}(X_{src}^{swap}), E_s^{tgt}(X_{tgt}^{swap}))$.

\begin{figure}[!t]
	\centerline{\includegraphics[width=0.9\columnwidth]{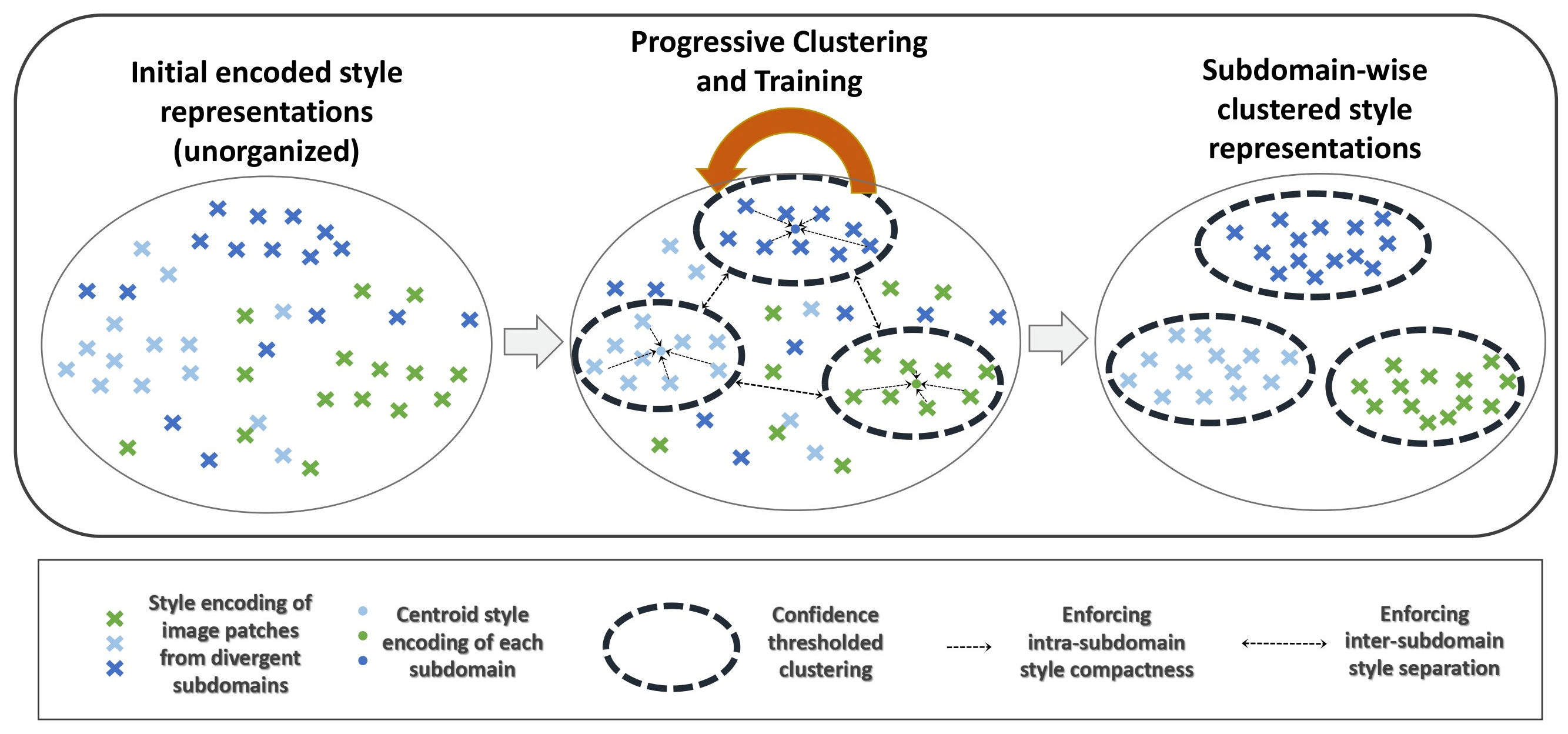}}
	\caption{Illustration of the progressive clustering and separation strategy. In this module, we enforce intra-subdomain style compactness as well as inter-subdomain style separation to benefit feature disentanglement. Considering that the pseudo subdomain labels are highly noisy, especially in the early training stage, we only compute losses based on reliable samples which have high confidence for clustering results. As the style encoder is gradually trained, more samples will become reliable and consequently the style encodings of all image patches will form clear cluster organizations.
	}
	\label{fig:pcs}
\end{figure}
\subsubsection{Progressive clustering and separation of cancer-specific subdomains}
\label{sec:kmeans}
With respect to the drastic distribution inner-variance present in the target domain, a severe defect of DRIT is that it priorly assumes the target domain to be homogeneous and expects the encoded style attributes for all images to follow a uniform distribution. 
In this case, cancer-incurred style variations would be neglected, leading to distorted feature disentanglement and cross-domain alignment\;\citep{Park2020}.
To transcend the bottleneck, we introduce a progressive clustering and partitioning strategy for style feature regularization to explicitly explore the inner-discrepancy of histopathology images, as illustrated in Fig.\;\ref{fig:pcs}.

For images from the compound target domain, they can be categorized into $K$ subdomains based on their disentangled style feature vectors, which represent the cancer type-specific and image patch-wise low-level texture characteristics. Here, $K$ is a hyper-parameter indicating the number of different subdomains for all the patches in the target domain, which is related yet in practice often not equal to the number of different cancer types within the target domain due to the sampling variations across patches.
%To take advantage of this prior knowledge, 
We propose to first collect the subdomain-specific attributes of each target image modeled by the style encoder into a memory bank and employ the K-means clustering technique\;\citep{Kanungo2002efficient} on top to secure the centroid of each subdomain.
Here, the content of memory bank and the clustering results are concurrently updated with the progress of training.
Then, for well-grounded subdomain structuring and separation, we propose to enforce the inter-subpartition disparity and intra-subpartition affinity of the style encodings. 
Specifically, we encourage the style representations of instances in the target domain to be adjacent to the centroid of its subdomain and distant from the centroids of others.
%Previous works hereafter assign pseudo subdomain labels for all target images and simplify OCDA into multi-target DA, while we argue that such design is suboptimal since feature disentanglement is drastically biased in the initial stage of training. 
Thereafter, we propose to progressively select reliable samples for style encoder optimization and dynamically update the style centroids.
At the early stage of model training, most subdomain splits are noisy and the corresponding image samples are therefore excluded from optimization.
As the training progresses, the proportion of incorrect subpartition assignments could be decreased. 
The samples with high inference confidence are then incorporated in the training and further benefit the style encoding performance.
In practice, we compute Equation (1) to measure the style similarity between an image and its subdomain centroid as the clustering confidence metric and Equation (2) to attain the loss value: 
\begin{equation}S_{ij}=\frac{1}{\sum\limits_{k=1}^K \left( \frac{\left \| {z_s}_i^{img}-c_j^{img} \right \|}{\left \| {z_s}_i^{img}-c_k^{img} \right \|} \right) ^ \frac{2}{m-1}} ,\end{equation}
\begin{equation}\begin{split}
	\mathcal{L}_{style-cluster}^{target}&=\frac{1}{N^{img}}\sum\limits_{i=1}^{N^{img}} \Big[\left( S_{ij}>\gamma \right) \cdot \big(\left \| {z_s}_i^{img}-c_j^{img} \right \| \\& -\frac{1}{K-1}\sum\limits_{k=1,k \neq j}^K\left \| {z_s}_i^{img}-c_k^{img} \right \| \big)\Big].
\end{split} \end{equation}
Here, given an image instance $i$ and a subdomain index $j$, ${z_s}_i^{img}$ and $c_{j}^{img}$ denote the style encoding of the instance and the centroid of the subdomain, respectively.
$K$ and $N^{img}$ denote the number of target subdomains and image samples in total. 
We conduct thorough analysis on the effect of different latent subdomain numbers in the following sections.
$m$ is a parameter to regulate the fuzziness of the measurement and set to 2 for $l_2$ normalization.
$\gamma$ corresponds to the confidence threshold that image tiles with confidence lower than $\gamma$ are considered as spurious and would be excluded from model training.
%More details can be found in the appendix.

\subsubsection{Shape and structure preservation along image translation}
Image translation techniques based on generative neural network model\;(e.g., CycleGAN and DRIT) have shown remarkable success in the histopathology domain to handle image appearance variations caused by discrepant image modalities and staining techniques\;\citep{Liu2020pdam, De2021residual}. 
However, due to the lack of supervision to explicitly induce shape and structure consistency, the nuclei in synthesized images suffer from severe deformation which inevitably results in the misalignment between synthesized images and instance segmentation labels (showcased in Section\;\ref{sec:effect_of_shape_preserve}).

To this end, we propose to set up two auxiliary blocks on top of the image translation pipeline for precise preservation of nucleus shape and structural details.
The synthesized target-like source image is forwarded to both of the branches in parallel. In the semantic segmentation branch, we employ a RGB-space feature encoder followed by a binary mask predictor to separate nuclei regions from the background. As for the boundary delineation branch, we first transform the input image from RGB color space to HED color space and extract H-space color map for further processing, so as to utilize the unique characteristics of H\&E stained histopathology images and highlight the nuclei boundaries. Thereafter, a feature encoder and an object boundary predictor are similarly utilized for nuclei boundary prediction. By enforcing the semantic masks and the boundaries of the synthesis images to be consistent with those in the raw image, nucleus over-generation and shape deformation along image translation can be effectively mitigated.
%Refer to 21-Panoptic Feature Fusion to see how to summarize all losses in one section.
%Cross entropy loss is used to supervise the consistency of mask and boundary, respectively defined as:
%\begin{equation}\mathcal{L}_{mask}^{fake-tgt}=-\frac{1}{N}\sum_{i=1}^N \left( y_i^m \cdot \log{x_i^m} \\ + \left(1-y_i^m\right) \cdot \log{\left(1-x_i^m\right)} \right)\end{equation}

\subsubsection{Overall training objectives}
As demonstrated in Fig.\;\ref{fig:stage1}, the overall loss function for Stage I cross-domain image alignment is composed of several items:
\begin{equation} \begin{split}
	\mathcal{L}_{DRIT}=&\mathcal{L}_{adv}^{content}+\mathcal{L}_{image-adv}^{source}+\mathcal{L}_{image-adv}^{target}\\&+\mathcal{L}_{cycle}^{source}+\mathcal{L}_{cycle}^{target}
	+\mathcal{L}_{style-reg}^{source}+\mathcal{L}_{style-reg}^{target} ,
\end{split} \end{equation}
\begin{equation}\begin{split}
	\mathcal{L}_{stageI}=&\lambda_1 \mathcal{L}_{DRIT}+\lambda_2 \mathcal{L}_{style-cluster}^{target}\\
	&+\lambda_3 \mathcal{L}_{mask}^{fake-target}+\lambda_4 \mathcal{L}_{boundary}^{fake-target} ,
\end{split} \end{equation}
where $\lambda$ represents hyper-parameter to control weights of each module. $\mathcal{L}_{DRIT}$ includes the typical adversarial and image reconstruction losses designed in DRIT, along with the regularization losses to constrain the style vectors can be drawn from a prior Gaussian distribution $N(0,1)$. $\mathcal{L}_{mask}^{fake-target}$ and $\mathcal{L}_{mask}^{fake-target}$ denote the cross entropy loss to supervise mask and boundary prediction respectively. 

\subsubsection{Style randomization for diverse image synthesis}
At the end of Stage I, we leverage the trained image translation model to generate target-like source images and hence mitigate image-level domain shifts. 
Motivated by the discovery that style augmentation considerably improves the model's generalization capability\;\citep{Jackson2019style}, we adopt the style randomization technique for diverse and arbitrary histopathology-like image synthesis.
Specifically, instead of synthesizing images conditioned on style features extracted from real histopathology image patches, arbitrary style attribute vectors are sampled from a prior Gaussian distribution $N(0,1)$ and integrated with content features of source image patches for cross-domain image translation.
The aim is to learn domain-invariant visual representations via the augmented images and consequently alleviate the performance drop of the trained model on open testing subdomains.

\begin{figure}[!t]
	\centerline{\includegraphics[width=1.0\columnwidth]{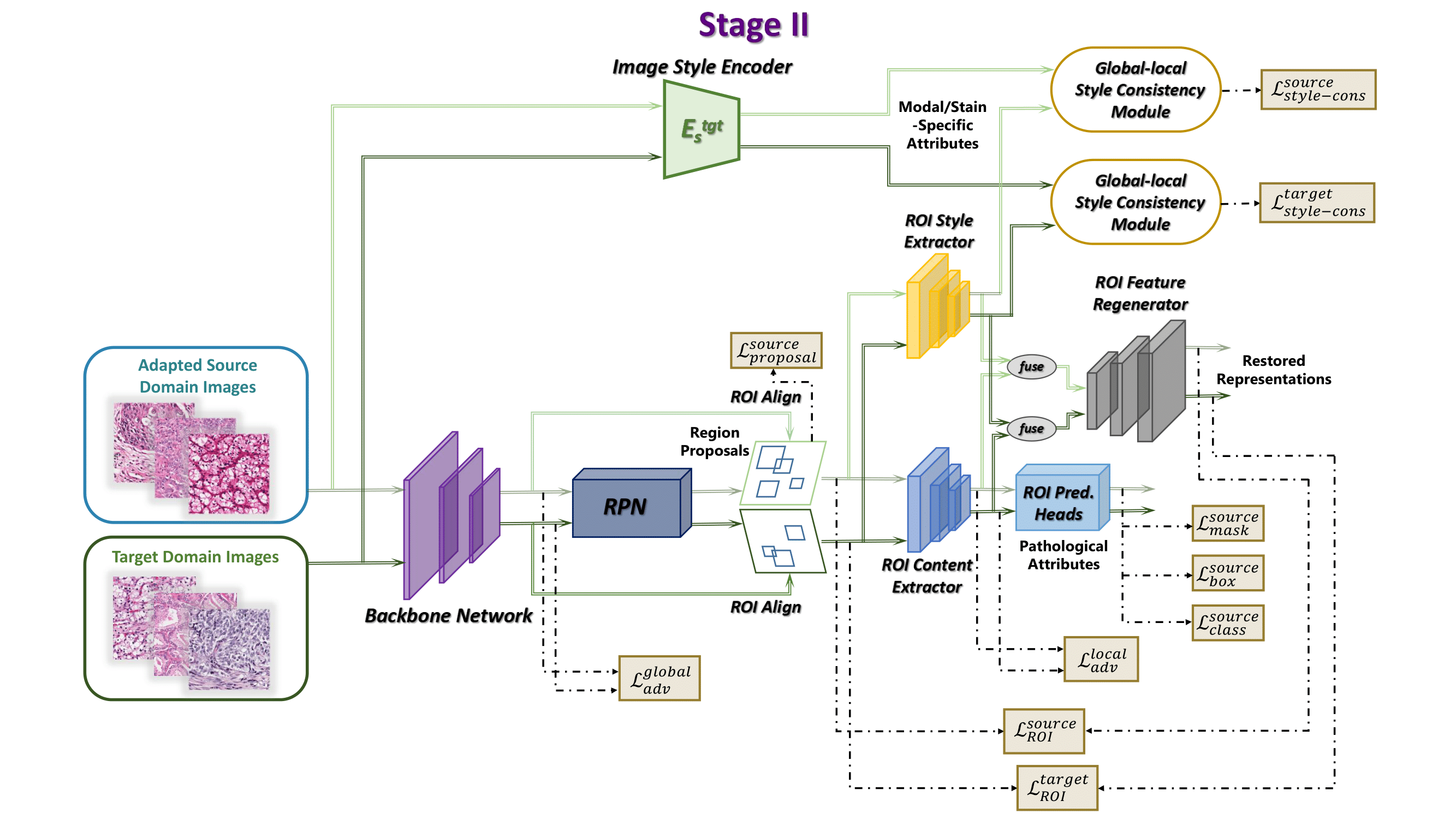}}
	\caption{Overview of Stage II for the proposed two-stage framework. In this stage, the inputs are the target-like source images synthesized with model trained in Stage I. Afterward, cross-domain feature alignment is performed via a Mask RCNN-based instance-level feature disentanglement network for domain adaptive instance segmentation.}
	\label{fig:stage2}
\end{figure}
\subsection{Stage II: Cross-domain feature alignment with local instance-level disentanglement}
\subsubsection{Backbone}
Following the design of previous works\;\citep{Liu2020pdam, Hsu2021}, we build the Stage II model upon the commonly adopted Mask R-CNN\;\citep{He2017mask} architecture but propose several modifications to achieve instance-level feature disentanglement and consequently alleviate the domain-specific factors within the feature representations for detection and segmentation task learning. 
%Specifically, two feature extractors are deployed in parallel for simultaneous ROI content and style encoding, followed by a feature regenerator to re-fuse the disentangled repensentations with a ROI feature consistency constraint to circumvent potential information loss in the feature encoding and disentanglement step.
All modules are shared for images from the target-like source domain and real target domain.
The detailed framework of Stage II model can be referred to in Fig.\;\ref{fig:stage2}.

\subsubsection{Domain-invariant feature alignment via feature disentanglement}
In the domain adaptive Mask R-CNN framework, feature representations forwarded to region proposal network (RPN) and ROI prediction heads are expected to be agnostic and indistinguishable across domains so as to prevent the model from overfitting to source domain data.
To this end, two feature extractors are deployed in parallel for simultaneous ROI content and style encoding, followed by a feature regenerator to re-fuse the disentangled representations with a ROI feature consistency constraint $L_{ROI}$ to circumvent potential information loss in the feature encoding and disentanglement step.
Moreover, two adversarial domain discriminators are deployed for the global features extracted by the backbone network and the disentangled instance-level content features respectively. Gradient reversal layers (GRLs) are inserted ahead of the discriminators to incorporate their training along with the main body.

\subsubsection{Global-local style consistency}
\label{sec:method:global-local_consis}
It is noted that for feature disentanglement model, the domain-invariant content and domain-specific style attributes are mutual-complementary and orthogonal in nature\;\citep{Wu2021vector}.
In this regard, precise and distinctive modeling of subdomain-specific characteristics plays a vital role for unbiased encoding of content attributes and is a crucial intermediate step towards ideal disentanglement on instance-level features, especially in the context of OCDA.
To regularize the encoded ROI style representation, a straightforward approach is to adopt the similar mechanism employed in Stage I, which is to at first assign subdomain labels for each instance based on its style representation and then encourage intra-subdomain style compactness as well as inter-subdomain style separation.
However, given the multi-class nature of cells, nuclei from the same images but of different categories typically possess divergent shape and spatial distribution. We find in practice that such cross-category heterogeneity inside each subdomain inevitably incurs serious style inconsistency and drastically compromises the accuracy of the assigned instance-level subdomain labels based on clustering.

As a result, we design a global-local style consistency mechanism to attain stable and category-agnostic instance-level style representation. In order to restrict the encoded style attribute to focus on subdomain-specific characteristics and exclude pattern variations caused by different nucleus categories, we assign a subdomain label for each local instance based on the global image-level style representation.
In details, the image-level style encoder trained in Stage I is reused to extract global style representation for images from the entire domain and then K-means clustering is applied on top to assign a subdomain label for each image.
It is noted that the clustering results obtained here are different from the ones in Stage I, as additional images synthesized with style randomization are further integrated.
Next, we enforce all instances from images with the same subdomain label to share close style representations and instances from images with different subdomain labels to possess disparate ones. A style consistency loss is designed for this objective:
\begin{equation}\begin{split}
	\mathcal{L}_{style-cons}&=\frac{1}{N^{ins}}\sum\limits_{i=1}^{N^{ins}} \big(\left \| {z_s}_i^{ins}-c_j^{ins} \right \| \\& -\frac{1}{K-1}\sum\limits_{k=1,k \neq j}^K\left \| {z_s}_i^{ins}-c_k^{ins} \right \| \big) ,
\end{split} \end{equation}
where ${z_s}_i^{ins}$ and $c_j^{ins}$ respectively denote the style encoding of a ROI instance and its subdomain centroid. 
$c_j^{ins}$ is attained similar to Stage I that for each subdomain, 
a memory bank of instance-level style feature vectors is firstly maintained. 
Then, the centroid for each subdomain is calculated by averaging all instance style encodings from images with the corresponding subdomain label.
$K$ and $N^{ins}$ denote the number of target subdomains and ROI instances in total.

\subsubsection{Overall training objectives}
To summarize, the overall training objective of Stage II is to minimize the following losses:
\begin{equation}\begin{split}
	\mathcal{L}&_{stageII}=\lambda_5 \mathcal{L}_{MaskRCNN}^{source}+\lambda_6 (\mathcal{L}_{adv}^{global}+\mathcal{L}_{adv}^{local}) \\& +\lambda_7 (\mathcal{L}_{ROI}^{source}+\mathcal{L}_{ROI}^{target})+
	\lambda_8 (\mathcal{L}_{style-cons}^{source}+\mathcal{L}_{style-cons}^{target}) ,
\end{split} \end{equation}
where $\mathcal{L}_{MaskRCNN}^{source}$ is the standard Mask R-CNN instance segmentation loss for images from the source domain and can be formulated as:
\begin{equation}\begin{split}
	%\mathcal{L}_{MaskRCNN}^{source}=&\mathcal{L}_{proposal}^{source}+\mathcal{L}_{mask}^{source} \\& +\mathcal{L}_{box}^{source}+\mathcal{L}_{class}^{source}
	\mathcal{L}_{MaskRCNN}^{source}=\mathcal{L}_{proposal}^{source}+\mathcal{L}_{mask}^{source} +\mathcal{L}_{box}^{source}+\mathcal{L}_{class}^{source} .
\end{split} \end{equation}
$\mathcal{L}_{adv}^{global}$ and $\mathcal{L}_{adv}^{local}$ represent the cross entropy losses for domain discriminators at two levels. $(\mathcal{L}_{ROI}^{source} + \mathcal{L}_{ROI}^{target})$ and $(\mathcal{L}_{style-cons}^{source} + \mathcal{L}_{style-cons}^{target})$ respectively denote the $L1$ reconstruction loss of ROI features and global-local style consistency loss for both the synthesized target-like source domain and the real target domain.

\section{Experiments}
\subsection{Datasets}
To comprehensively verify our method, we consider two representative cross-domain nuclei instance segmentation scenarios, cross-modality and cross-stain. 
\subsubsection{Cross-modality adaptation}\label{sec:exp_datasplit_modality}
Towards straighforward comparison against previous works focusing on cross-modality adaptation, we choose the fluorescence microscopy BBBC039 dataset\;\citep{Ljosa2012annotated} as the source domain and two histopathology image datasets containing H\&E-stained images of multiple cancer types, namely Kumar\;\citep{Kumar2017} and CPM17\;\citep{Vu2019methods}, are employed as the target domain.

\begin{figure}[!t]
	\centerline{\includegraphics[width=\columnwidth]{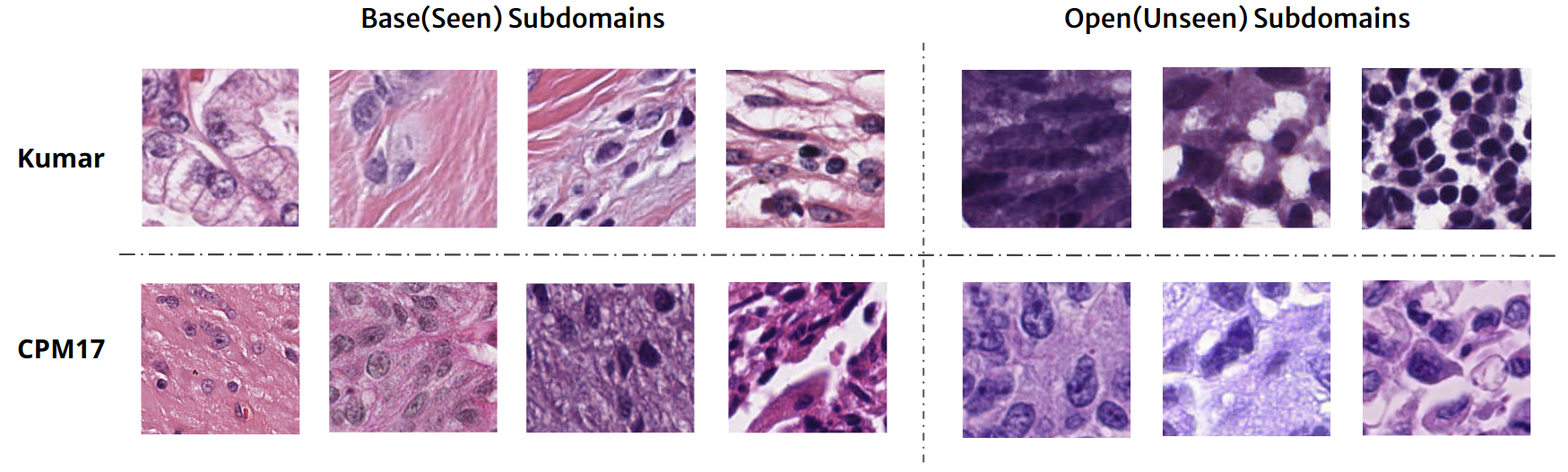}}
	\caption{Example image tiles from the base and open subdomains for two histopathology data collections. It can be observed that data distribution shifts evidently exist across the two sets of subdomains.}
	\label{fig:example}
\end{figure}

Following the same data split and preprocessing procedure as the previous work\;\citep{Liu2020pdam}, with the BBBC039 dataset, 100 training images and 50 validation images are utilized. 
With common data augmentation techniques including scaling, rotation, and flipping employed, around 10000 image patches in size 256 $\times$ 256 are extracted from the training set. 
As for the multi-cancer histopathology image datasets, to fit in the OCDA setting and evaluate the model's generalizability on unseen subdomains, images of specific cancer types are excluded from the training set and only exist in the testing set. 
In the Kumar dataset, which contains in total 30 1000 $\times$ 1000 histopathology images from seven types of cancer, 16 images from liver, kidney, prostate, and breast cancer\;(four images from each type of cancer) are selected to be mixed as the compound target domain and formulated into the training set, while the remaining 14 images
%(2 images from each type of cancer) 
comprising three unseen subdomains\;(bladder, stomach, colon) are left for testing. 
The data split strategy is suggested to ensure that evident visual discrepancy and data distribution shifts exist across the base and open subdomains, as illustrated in Fig.\;\ref{fig:example}.
With this respect, the generalizability of the trained model to be adapted to the clinical wild can be fairly evaluated.
Likewise, in the CPM17 dataset consisting of 64 500 $\times$ 500 or 600 $\times$ 600 images from four types of cancer, images from lower grade glioma (LGG) cancer are formulated as the open subdomain for generalization assessment. 
24 images from non small cell lung cancer (NSCLC), head and neck squamous cell carcinoma (HNSCC), and glioblastoma multiforme (GBM) cancer (eight images from each type of cancer) compose the training set, and 32 images (eight images from each type of cancer) are used for testing, as summarized in Table\;\ref{tab:data_split}. 
All the images are randomly cropped into 256 $\times$ 256 patches during preprocessing.

\subsubsection{Cross-stain adaptation}\label{sec:exp_datasplit_stain}
In this scenario, we aim to adapt knowledge learned from IHC-stained images to H\&E-stained images. DataSeg\;\citep{Shu2020marker} contains 52 images in size 200 $\times$ 200 which are captured at a core center or a region center surrounded by a large number of positively stained nuclei. After scaling and stitching, 256 $\times$ 256 image patches are generated and formulate the source domain.
With respect to target H\&E-stained domain, we reuse the Kumar dataset and follow the same data split strategy as mentioned in Section\;\ref{sec:exp_datasplit_modality}.

\begin{table}[!t]
\centering
\caption{Data split strategy for histopathology image datasets. Numbers in brackets denote the number of images for each type of cancer. Unseen testing subdomains are highlighted in \textbf{bold}.}
\begin{adjustbox}{width=0.8\columnwidth,center}
\begin{tabular}{c|c|c}
	\toprule
	Dataset & Training Set & Testing Set\\ 
	\midrule
	\multirow{3}{*}{\small Kumar} & \multirow{2}{*}{\small Liver(4) Kideney(4)} & \small Liver(2) Kideney(2) \\
	& \multirow{2}{*}{\small Prostate(4) Breast(4)} & \small Prostate(2) Breast(2) \\
	& & \small \bf{Bladder(2) Stomach(2) Colon(2)} \\ 
	\midrule
	\multirow{2}{*}{\small CPM17} & \small GBM(8) NSCLC(8) & \small GBM(8) NSCLC(8) \\
	& \small HNSCC(8) & \small HNSCC(8) \bf{LGG(8)} \\
	\bottomrule
\end{tabular}
\end{adjustbox}
\label{tab:data_split}
\end{table}

\subsection{Implementation details}
\label{sec:exp_imple_detail}
In Stage I framework, all the content encoders, style encoders, image generators, feature discriminators, and image discriminators are implemented with the same architecture as in \cite{Lee2020drit}. 
The implementation for both mask and boundary prediction branches follows the structure of the semantic segmentation branch in \cite{Kirillov2019panoptic} to fuse the multi-scale features and then conduct prediction.
In Stage II, we employ the Mask R-CNN with ResNet101\;\citep{He2016deep} in conjunction with FPN\;\citep{Lin2017feature} as the backbone.

We implement the proposed method with Pytorch and MaskRCNN-Benchmark\;\citep{Massa2018maskrcnn}. As for hyperparameter configurations, we empirically set the confidence threshold $\gamma = 0.5$ in Equation (2), $N^{mem} = 10^4, l^{style} = 32, F^{upd} = 500, T = 10^5$ in Algorithm 1, $\lambda_1 = 1, \lambda_2 = 2, \lambda_3 = 5, \lambda_4 = 10$ in Equation (4), $\lambda_5 = 1, \lambda_6 = 1, \lambda_7 = 2, \lambda_8 = 1$ in Equation (6). 
Adam optimizer with a learning-rate of $10^{-4}$ is employed to train Stage I's model, whereas following the design of \cite{Massa2018maskrcnn}, SGD optimizer with an initial learning-rate of $5 \times 10^{-4}$ is used in Stage II's training.

\subsection{Evaluation metrics}
For the purpose of fair comparison, we adopt three metrics to evaluate the performance of nuclei instance segmentation which are broadly used in previous works. 
Panoptic quality\;(PQ) is a unified score which integrates detection quality\;(DQ) and segmentation\;(SQ)\;\citep{Graham2019hover}.
The consolidated metric inherits capability to simultaneously measure the accuracy with respect to both detection and segmentation tasks.
It is considered as a robust quantification for comprehensive evaluation of instance segmentation result. 
% and can be further splitted into detection quality (DQ) and segmentation quality (SQ) for better interpretability
Additionally, we adopt DICE and AJI\;\citep{Kumar2017} to perform supplemental evaluation at semantic and instance level respectively.

\renewcommand{\arraystretch}{1.3} %控制行高
\begin{table}[!t]
\centering
\fontsize{6}{7}\selectfont
\begin{threeparttable}
	\caption{Performance comparison for cross-modality nuclei instance segmentation on the BBBC039 $\rightarrow$ Kumar benchmark. The best results among UDA and OCDA approaches are highlighted in \textbf{bold}.}
	\label{tab:result_modality_kumar}
	\setlength{\tabcolsep}{1.mm}{
		\begin{tabular}{c|ccccc}
			\toprule
			\multirow{2}{*}{Method}&
			\multicolumn{5}{c}{BBBC039 $\rightarrow$ Kumar}\cr
			\cmidrule(lr){2-6}
			&DICE&AJI&DQ&SQ&PQ\cr
			\midrule				
			Source-only&$0.4330\pm0.1926$&$0.2817\pm0.1731$&$0.6140\pm0.2053$&$0.5882\pm0.1170$&$0.3145\pm0.1872$\cr				
			DARCNN\;\citep{Hsu2021}&$0.6619\pm 0.0785$&$0.4461\pm0.1425$&$0.6877\pm0.1360$&$0.6253\pm0.0814$&$0.4193\pm0.1704$\cr
			PDAM\;\citep{Liu2020pdam}&$0.7904\pm0.0474$&$0.5653\pm0.0751$&$0.7154\pm0.1055$&$0.7180\pm0.0626$&$0.5249\pm0.0884$\cr
			DHA\;\citep{Park2020}&$0.7505\pm0.0592$&$0.4958\pm0.1176$&$0.6907\pm0.1298$&$0.6512\pm0.0860$&$0.4473\pm0.1492$\cr
			CSFU\;\citep{Gong2021cluster}&$0.6983\pm0.0767$&$0.4257\pm0.1484$&$0.6552\pm0.1495$&$0.6544\pm0.0702$&$0.4223\pm0.1362$\cr
			ML-BPM\;\citep{pan2022ml}&$0.7120\pm0.0675$&$0.4715\pm0.1301$&$0.6861\pm0.1181$&$0.6697\pm0.0619$&$0.4638\pm0.1093$\cr
			Ours&$\bf{0.7930\pm0.0446}$&$\bf{0.5797\pm0.0740}$&$\bf{0.7466\pm0.0799}$&$\bf{0.7373\pm0.0307}$&$\bf{0.5527\pm0.0795}$\cr
			Supervised&$0.7886\pm0.0531$&$0.5735\pm0.0855$&$0.7248\pm0.0661$&$0.7307\pm0.0408$&$0.5388\pm0.1004$\cr
			\bottomrule
	\end{tabular}}
\end{threeparttable}
\end{table}

\begin{table}[!t]
\centering
\fontsize{6}{7}\selectfont
\begin{threeparttable}
	\caption{Performance comparison for cross-modality nuclei instance segmentation on the BBBC039 $\rightarrow$ CPM17 benchmark. The best results among UDA and OCDA approaches are highlighted in \textbf{bold}.}
	\label{tab:result_modality_cpm}
	\setlength{\tabcolsep}{1mm}{
		\begin{tabular}{c|ccccc}
			\toprule
			\multirow{2}{*}{Method}&\multicolumn{5}{c}{BBBC039 $\rightarrow$ CPM17}\cr
			\cmidrule(lr){2-6}
			&DICE&AJI&DQ&SQ&PQ\cr
			\midrule				
			Source-only&$0.5156\pm0.1565$&$ 0.3340\pm0.1384$&$0.6250\pm0.1606$&$0.5479\pm0.0941$&$0.3408\pm0.1456$\cr
			DARCNN\;\citep{Hsu2021}&$0.7155\pm0.0662$&$0.5028\pm0.0921$&$0.6765\pm0.1261$&$0.6601\pm0.0759$&$0.4494\pm0.1092$\cr
			PDAM\;\citep{Liu2020pdam}&$0.8110\pm0.0445$&$0.5913\pm0.0881$&$0.7785\pm0.0955$&$0.7236\pm0.0546$&$0.5638\pm0.0962$\cr
			DHA\;\citep{Park2020}&$0.7954\pm0.0488$&$0.5560\pm0.0961$&$0.7410\pm0.1031$&$0.7096\pm0.0581$&$0.5202\pm0.1074$\cr
			CSFU\;\citep{Gong2021cluster}&$0.6909\pm0.0849$&$0.5110\pm0.1194$&$0.7162\pm0.1187$&$0.6630\pm0.0925$&$0.4715\pm0.1140$\cr
			ML-BPM\;\citep{pan2022ml}&$0.7442\pm0.0537$&$0.5359\pm0.1042$&$0.7286\pm0.1109$&$0.6740\pm0.0747$&$0.4837\pm0.1107$\cr
			Ours&$\bf{0.8237\pm0.0471}$&$\bf{0.6090\pm0.0867}$&$\bf{0.8210\pm0.0849}$&$\bf{0.7624\pm0.0337}$&$\bf{0.6274\pm0.0867}$\cr
			Supervised&$0.8496\pm0.0418$&$0.6614\pm0.0857$&$0.8609\pm0.0683$&$0.7653\pm0.0294$&$0.6595\pm0.0797$\cr
			\bottomrule
	\end{tabular}}
\end{threeparttable}
\end{table}
\subsection{Evaluation on cross-modality adaptation}
We at first validate the proposed method on cross-modality domain adaptation, i.e., from fluorescence microscopy to
histopathology images. We compare against the state-of-the-art methods, including conventional UDA methods, OCDA method, and fully supervised method. The details are provided as follows:
\begin{itemize}
\item \textbf{UDA: DARCNN}\;\citep{Hsu2021} and \textbf{PDAM}\;\citep{Liu2020pdam}: DARCNN and PDAM are the most state-of-the-art domain adaptation methods for cross-modality nuclei instance segmentation. Advanced techniques such as self-supervised representation consistency loss and feature similarity maximization mechanism are exploited in these works. 
%Nevertheless, they consider histopathology image as a homogeneous domain and align the entire domain integrally. 
\textbf{OCDA: DHA}\;\citep{Park2020}, \textbf{CSFU}\;\citep{Gong2021cluster}, and \textbf{ML-BPM}\;\citep{pan2022ml}: DHA, CSFU, and ML-BPM are the recent efforts proposed for addressing the OCDA challenge in the context of semantic segmentation.
%is an OCDA framework which performs image clustering, synthesis, and adversarial feature alignment in sequence. 
We extend those methods to the instance segmentation task by replacing their semantic segmentation branch with a Mask R-CNN module.
\textbf{Supervised}: We train a Panoptic FPN\;\citep{Kirillov2019panoptic} model exploiting imaging data with high-quality annotations from both source and target in a fully-supervised manner to illustrate the performance upper bound of the nuclei analysis task.
Comparison methods in the context of UDA and OCDA are not required to compete with the supervised approach since they retain distinct levels of dependence on data and corresponding annotations.
\end{itemize}
%It is worth noting that, as elaborated in Section\;\ref{sec1}, assigning reliable subdomain labels to each histopathology image patch is infeasible. In this regard, we do not perform comparison with multi-target DA methods.

\begin{figure}[!t]
\centerline{\includegraphics[width=\columnwidth]{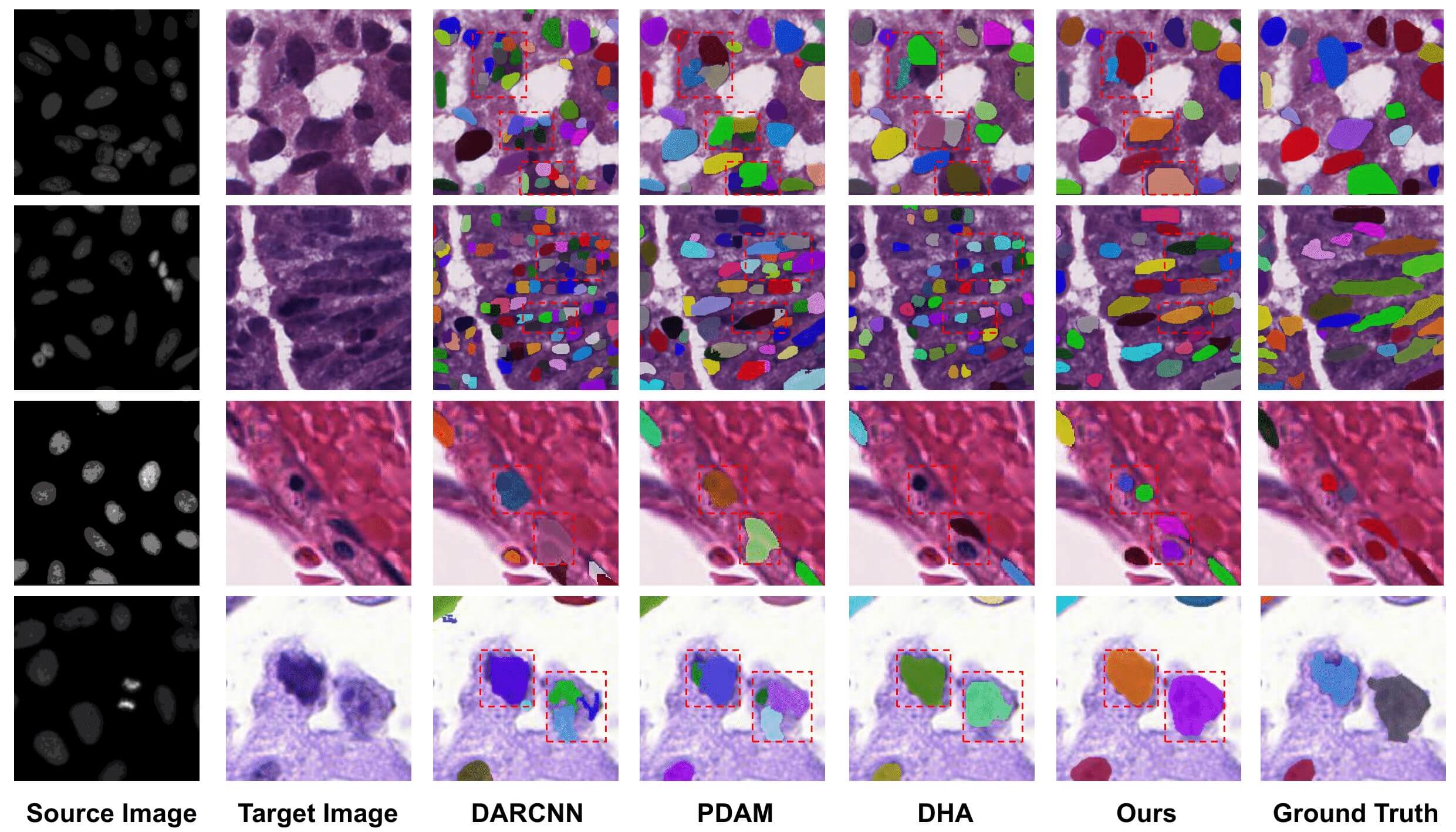}}
\caption{Visualization of cross-modality nuclei instance segmentation results on two H\&E-stained  histopathology datasets. The first column provides four examples of the source fluorescence microscopy images. The images of the top two rows are from Kumar dataset, and the bottom two rows are from CPM17 dataset. Separated nuclei are indicated with different colors. The red rectangles are plotted to highlight the difference of all results.}
\label{fig:source_fluo}
\end{figure}

The quantitative comparison results are presented in Table\;\ref{tab:result_modality_kumar} and \ref{tab:result_modality_cpm}, from which we can find that our method outperforms all the DA methods in all metrics. It even demonstrates superior performance compared with the fully supervised method in the Kumar dataset. 
We also perform one-tailed paired t-test to evaluate the
statistical significance between our proposed method and the competing UDA and OCDA methods in terms of PQ. All the resulting p-values are under 0.001, except when comparing with PDAM\;\citep{Liu2020pdam} on the BBBC039 to Kumar benchmark. However, the p-value in this case is still under 0.01. Those results indicate the statistical significance of our achieved improvements.
In particular, compared with DARCNN which only conducts feature-level alignment and target domain-specific model fine-tuning, our method suggests holistic cross-domain alignment and in consequence dramatically lifts the performance with the proposed two-stage disentanglement framework. 
PDAM adopts the similar two-stage model design yet solely proposes to obtain domain-agnostic representation for downstream tasks and neglects the multi-modal data distribution in histopathology images. In contrast, our method models the intra-domain heterogeneity of histopathology images explicitly by exploiting the rich subdomain-specific characteristics in both image translation and feature alignment. We further design a nucleus shape and structure preserving module to enhance the correspondance between the synthesized nuclei objects and the annotations. As a result, we exceed PDAM in both Kumar and CPM17 datasets, especially in terms of PQ on which we can observe an improvement over $3\%\sim6\%$.
DHA, CSFU, and ML-BPM are a set of frameworks specifically designed for OCDA which propose to tackle the intro-domain heterogeneity by firstly performing latent domain discovery and subsequently simplifying the OCDA setting into multi-target DA. 
Despite their success on global-level semantic segmentation, we observe that such approaches generally fail to capture the tissue/cancer-wise pattern variations in the histopathology domain and tend to generate erroneous subdomain labels, which inevitably incurs error accumulation along following steps. 
In addition, those frameworks only perform semantic-level adaptation, whereas our proposed method also benefits from instance-level adaptation via debiased representation decomposition. 
As shown in the quantitative comparison, our method attains considerably better results over all adaptation benchmarks.
%Given the previous observation, we instead propose a more flexible approach to handle the compound target domain with a style clustering loss and the progressive training strategy. Images with noisy style representations are excluded from the model training to avoid their side effects. 
%
In comparison with the Supervised Upper Bound, it is observed that our method even achieves superior results on the BBBC $\rightarrow$ Kumar benchmark, without any requirements for annotated target domain data.
Our method also attains appealing accuracy comparable to the supervised upper bound on the other adaptation benchmarks, which substantiates its promise in data-efficient scenarios.

%plot some result
Aside from the quantitative comparison, we additionally present the qualitative comparison results on four image patches of different cancer types in Fig.\;\ref{fig:source_fluo}. As can be seen from the red rectangles, all the competing methods seriously suffer from poor modeling of instance-level characteristics. To be specific, they either generate overdense instance predictions that one integrated nucleus is splitted into several isolated ones, or cannot precisely separate touching nuclei clusters. 
This observation provides an explaination for what we find in the quantitative comparison results that compared with AJI and Pixel-F1, our method surpasses the others in terms of PQ by a larger margin. 
AJI and Pixel-F1 only measure the overall score of instance segmentation and mainly focus on the prediction accuracy in pixel-level. In contrast, PQ is calculated by multiplying two instance-level metrics, detection quality (DQ), and segmentation quality (SQ). In other words, it is mostly determined by prediction accuracy of each individual object, rather than global pixel-wise accuracy.
Owing to its capability to precisely characterize instance-level and object-specific attributes, our method improves the previous works in terms of PQ remarkably.
%Comparison with DHA
Meanwhile, in particular to DHA, the prediction results indicate its inferior performance as it fails to detect and accurately delineate the boundaries of several nuclei. This can be attributed to its biased target domain partition and subdomain-wise image translation. In practice we find that in DHA, as the assigned subdomain labels are drastically noisy, the divided subdomains are still mixed and with multi-modal data distribution. As a consequence, the following subdomain-wise image translation step is prone to synthesizing unrealistic and spurious nucleus texture patterns, which leads to failure in cross-domain nuclei instance segmentation.

\begin{figure}[!t]
\centerline{\includegraphics[width=\columnwidth]{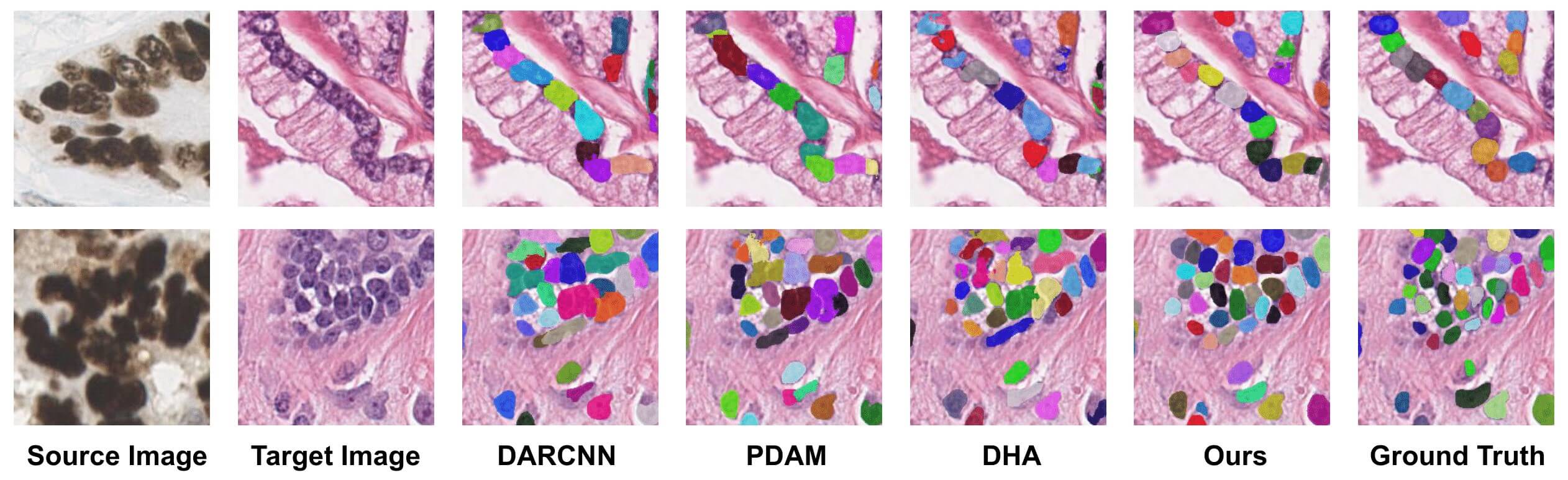}}
\caption{Visualization of cross-stain nuclei instance segmentation results on Kumar dataset. The first column provides two examples of the source IHC-stained images.}
\label{fig:source_IHC}
\end{figure}
\begin{table}[!t]
\centering
\fontsize{7}{8}\selectfont
\begin{threeparttable}
	\caption{Performance comparison in specific to seen subdomains and unseen subdomains on Kumar dataset. The best results are highlighted in \textbf{bold}.}
	\label{tab:seen_unseen_comparison}
	\setlength{\tabcolsep}{2.5mm}{
		\begin{tabular}{c|ccc}
			\toprule
			\multirow{2}{*}{Method}&
			\multicolumn{3}{c}{PQ}\cr
			\cmidrule(lr){2-4}
			&seen&unseen&all\cr
			\midrule
			Source-only&$0.3008$&$0.3329$&$0.3145$\cr
			DARCNN\;\citep{Hsu2021}&$0.4108$&$0.4306$&$0.4193$\cr
			PDAM\;\citep{Liu2020pdam}&$0.5173$&$0.5351$&$0.5249$\cr
			DHA\;\citep{Park2020}&$0.4434$&$0.4525$&$0.4473$\cr
			Ours&$0.5347$&$\bf{0.5767}$&$\bf{0.5527}$\cr
			Supervised Upper Bound&$\bf{0.5384}$&$0.5394$&$0.5388$\cr				
			\bottomrule
	\end{tabular}}
\end{threeparttable}
\label{tab:seen_unseen}
\end{table}

%Seen Unseen
To evaluate the generalization capability to unseen subdomains in the OCDA setting, we further present the quantitative results on testing images from seen subdomains and unseen subdomains individually.
As shown in Table\;\ref{tab:seen_unseen_comparison}, our method exhibits outstanding robustness to unseen subdomains. 
%Why outperform supervised methods
It is the reason why our method can outperform fully supervised upper bound in this dataset. 
Fully supervised method is vulnerable to distribution shifts among training set and testing set, which is relatively serious in Kumar. Since our method is built upon the disentanglement architecture, it is able to acquire domain-invariant representations at both the image level and the instance level. Along with the style randomization technique deployed for diverse data augmentation, our proposed method demonstrates effectiveness in dealing with images from unseen subdomains.

\subsection{Evaluation on cross-stain adaptation}
\begin{table}[!t]
\centering
\fontsize{6}{7}\selectfont
\begin{threeparttable}
	\caption{Performance comparison for cross-stain nuclei instance segmentation. The best results among UDA and OCDA approaches are highlighted in \textbf{bold}.}
	\label{tab:cross_stain_performance_comparison}
	\setlength{\tabcolsep}{1mm}{
		\begin{tabular}{c|ccccc}
			\toprule
			\multirow{2}{*}{Method}&
			\multicolumn{5}{c}{DataSeg(IHC) $\rightarrow$ Kumar}\cr
			\cmidrule(lr){2-6}
			&DICE&AJI&DQ&SQ&PQ\cr
			\midrule				
			Source-only&$0.6436\pm0.0693$&$0.4055\pm0.0895$&$0.5919\pm0.1624$&$0.6129\pm0.0755$&$0.3511\pm0.1051$\cr
			DARCNN\;\citep{Hsu2021}&$0.7344\pm0.0638$&$0.4365\pm0.0877$&$0.6208\pm0.1530$&$0.6103\pm0.0927$&$0.3828\pm0.1026$\cr
			PDAM\;\citep{Liu2020pdam}&$0.7515\pm0.0549$&$0.4663\pm0.0757$&$0.6455\pm0.1279$&$0.6492\pm0.0462$&$0.4136\pm0.1019$\cr
			DHA\;\citep{Park2020}&$0.7390\pm0.0594$&$0.4514\pm0.0833$&$0.6034\pm0.1383$&$0.6369\pm0.0553$&$0.3848\pm0.1048$\cr
			CSFU\;\citep{Gong2021cluster}&$0.7046\pm0.0822$&$0.4393\pm0.1064$&$0.6169\pm0.1600$&$0.6017\pm0.0861$&$0.3769\pm0.1210$\cr
			ML-BPM\;\citep{pan2022ml}&$0.7662\pm0.0498$&$0.4913\pm0.0726$&$0.6617\pm0.1012$&$0.6831\pm0.0705$&$0.4483\pm0.1070$\cr
			Ours&$\bf{0.7786\pm0.0461}$&$\bf{0.5472\pm0.0707}$&$\bf{0.6940\pm0.1118}$&$\bf{0.7197\pm0.0224}$&$\bf{0.5018\pm0.0911}$\cr
			\bottomrule
	\end{tabular}}
\end{threeparttable}
\end{table}
We further perform the comparison study on another cross-domain scenario, i.e. cross-stain adaptation, to verify the efficacy and robustness of our method. 
As mentioned in Section\;\ref{sec:exp_datasplit_stain}, IHC-stained histopathology image dataset DataSeg is set as the source domain, while H\&E-stained dataset Kumar is reused as the compound target domain.

The quantitative and qualitative comparison results are shown in Table\;\ref{tab:cross_stain_performance_comparison} and Fig.\;\ref{fig:source_IHC}, respectively. 
The overall observation is consistent with our previous discovery that our method reaches peak performance and even attains more significant improvements over the previous works. 
We postulate it is because the employed DataSeg dataset is rather small, and with the one-to-one image translation model (CycleGAN) adopted in previous works, only limited fake target images can be synthesized which consequently results in an intense over-fitting issue. In our method, due to the disentanglement framework and style randomization technique, a huge amount of images of various styles can be flexibly generated to bypass the obstacle. 
Moreover, we notice that since the IHC-stained histopathology images possess far more complicated texture patterns than the fluorescence microscopy ones, conventional image translation model without nucleus-specific supervision typically loses essential nucleus shape and structural details when conducting translation, which inevitably incurs mismatch between synthesized images and nuclei segmentation labels (showcased in Section\;\ref{sec:effect_of_shape_preserve}). In comparison, with the regularization of the proposed nucleus shape and structure preserving module, images generated by our method showcase promising semantic consistency and provide unbiased supervision for following steps.

\section{Discussion}
%\subsection{Effectiveness of xx module}
% In style clustering study, a t-sne-like figure can be plotted to compare the separation of subdomain before and after clustering loss
% In nucleus shape preservation study, refer to the PPT when meeting with shixiong.
% In style randomization study, present an example about how diverse images can be synthesized.
% In local-level disentanglement study, just provide the quantitative results with and without each module should be enough.
To explore and validate the effectiveness of the key modules deployed in our method, we conduct an ablation study on the BBBC039 to Kumar benchmark. The quantitative performance comparison is presented in Table\;\ref{tab:ablation_study}, where PCS denotes the progressive clustering and separation module, NSSP denotes the nucleus shape and structure preserving module, 
%SR denotes style randomization, 
ID denotes instance-level disentanglement, GLSC denotes the global-local style consistency module.
Additionally, to support our claim in Section\;\ref{sec:method:global-local_consis}, we provide the results when the clustering-based strategy in Stage I is similarly applied in Stage II for reference, denoted as Clust II.
\begin{table}[!t]
\centering
\fontsize{7}{8}\selectfont
\begin{threeparttable}
	\caption{Quantitative analysis of key components in our method on BBBC039 to Kumar benchmark. The best results are highlighted in \textbf{bold}.}
	\label{tab:ablation_study}
	\setlength{\tabcolsep}{2.5mm}{
		\begin{tabular}{c|ccc}
			\toprule
			\multirow{2}{*}{Method}&
			\multicolumn{3}{c}{BBBC039 $\rightarrow$ Kumar}\cr
			\cmidrule(lr){2-4}
			&DICE&AJI&PQ\cr
			\midrule				
			w/o PCS&$0.7857\pm0.0440$&$0.5605\pm0.0788$&$0.5318\pm0.0862$\cr
			w/o NSSP&$0.7410\pm0.0636$&$0.5015\pm0.0976$&$0.4558\pm0.1195$\cr
			%w/o SR&$0.5063\pm0.0853$&$0.5415\pm0.0709$&$0.7710\pm0.0492$\cr
			w/o ID&$0.7833\pm0.0449$&$0.5532\pm0.0918$&$0.5083\pm0.1052$\cr
			w/o GLSC&$0.7838\pm0.0470$&$0.5620\pm0.0822$&$0.5287\pm0.0865$\cr
			Clust II&$0.7855\pm0.0448$&$0.5541\pm0.0942$&$0.5074\pm0.1108$\cr
			DRIT\;\citep{Lee2020drit}&$0.7139\pm0.0701$&$0.4710\pm0.1368$&$0.3978\pm0.1560$\cr
			DRANet\;\citep{lee2021dranet}&$0.7198\pm0.0578$&$0.5007\pm0.1119$&$0.4153\pm0.1479$\cr
			Ours&$\bf{0.7930\pm0.0446}$&$\bf{0.5797\pm0.0740}$&$\bf{0.5527\pm0.0795}$\cr	
			\bottomrule
	\end{tabular}}
\end{threeparttable}
\end{table}
\subsection{Effectiveness of the progressive clustering and separation module}
\begin{figure}[!t]
\centerline{\includegraphics[width=0.8\columnwidth]{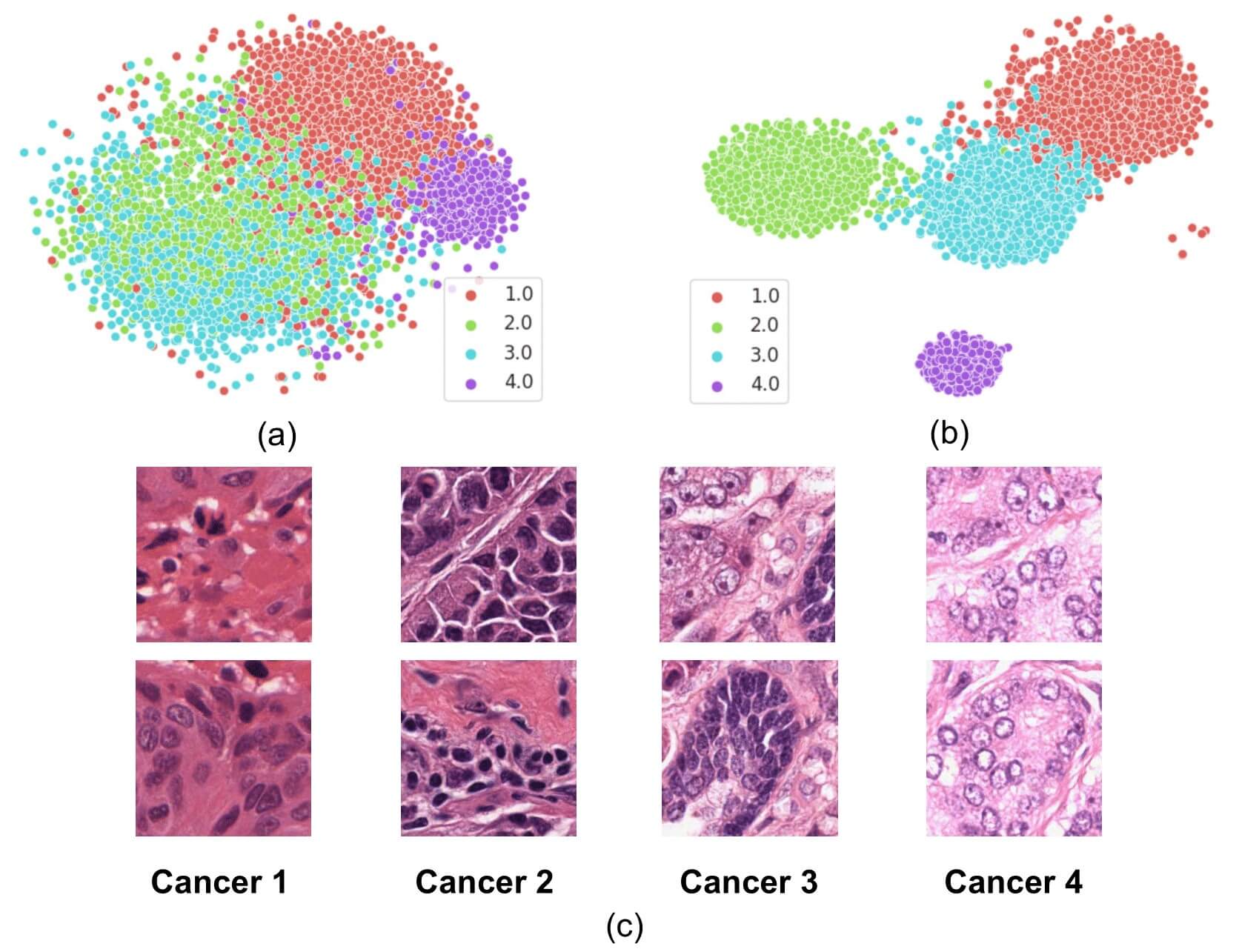}}
\caption{Visualization of clustering results. (a) t-SNE visualization of encoded style representations w/o PCS. (b) t-SNE visualization of encoded style representations with PCS. (c) Example images from each type of cancer.}
\label{fig:visual_clustering}
\end{figure}
We first evaluate the effectiveness of the progressive clustering and separation module in Stage I. To this end, we set $\lambda_2$ in Equation (4) to 0 and do not amplify the separation of subdomain-specific characteristics. In addition, we plot the clustering of the style encodings extracted from images of four divergent cancer types. As shown in Fig.\;\ref{fig:visual_clustering}, style representations encoded without the proposed module are highly mixed and indistinguishable among different cancer types, whereas with the regularization of the style separation module, style representations exhibit apparent cluster organization. The observation is supported by the quantitative comparison ($1^{st}$ row, Table\;\ref{tab:ablation_study}) as well as the fact that the instance segmentation results are enhanced under all three metrics with strengthened style division.

\subsection{Effect of different latent target subdomain numbers}
\begin{figure}[!t]
\centerline{\includegraphics[width=0.9\columnwidth]{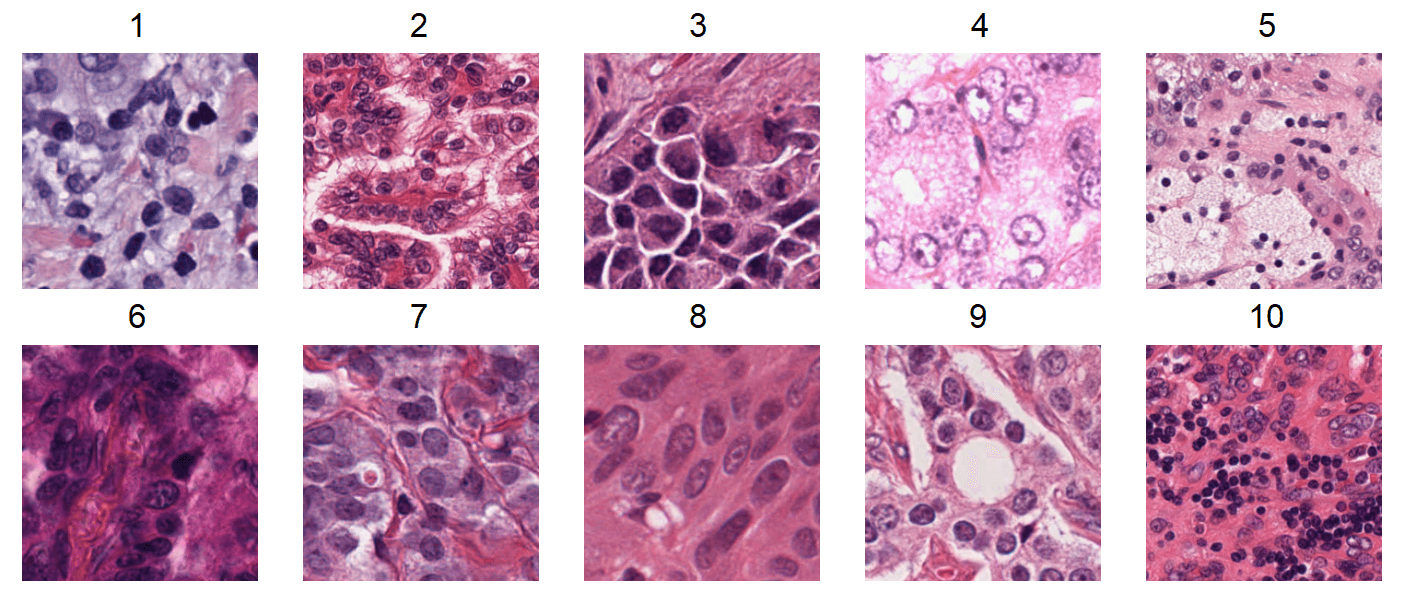}}
\caption{Example images of each clustered subdomain when $K$ is set to $10$.}
\label{fig:visual_subdomains}
\end{figure}

% Specifically, the effect of subdomain numbers K. We aim to illustrate that our method is not sensitive to this parameter and thus better than previous OCDA methods.
\begin{table}[!t]
\centering
\fontsize{7}{8}\selectfont
\begin{threeparttable}
	\caption{Quantitative parameter analysis of $K$ on the BBBC039 to Kumar benchmark. The best results are highlighted in \textbf{bold}.}
	\label{tab:parameter_K}
	\setlength{\tabcolsep}{2.5mm}{
		\begin{tabular}{c|ccc}
			\toprule
			\multirow{2}{*}{Setting}&
			\multicolumn{3}{c}{BBBC039 $\rightarrow$ Kumar}\cr
			\cmidrule(lr){2-4}
			&DICE&AJI&PQ\cr
			\midrule		
			w/o PCS&$0.7857\pm0.0440$&$0.5605\pm0.0788$&$0.5318\pm0.0862$\cr		
			$K=4$&$0.7867\pm0.0453$&$0.5639\pm0.0787$&$0.5366\pm0.0843$\cr
			$K=7$&$\bf{0.7943\pm0.0421}$&$0.5780\pm0.0729$&$0.5501\pm0.0821$\cr
			$K=10$&$0.7930\pm0.0446$&$\bf{0.5797\pm0.0740}$&$\bf{0.5527\pm0.0795}$\cr
			$K=13$&$0.7895\pm0.0452$&$0.5763\pm0.0795$&$0.5464\pm0.0817$\cr	
			$K=16$&$0.7882\pm0.0463$&$0.5746\pm0.0811$&$0.5468\pm0.0839$\cr
			\bottomrule
	\end{tabular}}
\end{threeparttable}
\end{table}
The number of latent target subdomains, $K$, is an important hyperparameter in the previous progressive clustering and separation module. We further study the effect of different choices of $K$. 
Specifically, considering that the training target domain images are from four types of cancer and the training image patches are cropped from 16 image tiles, we vary the number of $K$ from 4 to 16, with an interval of 3, and present the corresponding results in Table\;\ref{tab:parameter_K}.
It is observed that our method is robust to the hyperparameter $K$ such that it outperforms the w/o PCS baseline under all settings.
When $K=10$, our method achieves the highest performance in terms of PQ and AJI. Example images of each clustered subdomain are demonstrated in Fig.\;\ref{fig:visual_subdomains}.
It is noteworthy that although these images are only from four types of cancer, each image patch cluster contains its own distinctive style pattern and should be considered as an individual subdomain.
This finding supports our statement in Section\;\ref{sec1} that multi-target DA is unsuitable for our task where subdomain labels cannot be directly assigned according to the image's cancer type.
In comparison, as for the extreme case when $K$ is set to the number of cancer types, i.e. $K=4$, the performance is relatively modest, which indicates that the divergent style characteristics are not fully exploited. On the other hand, when $K$ is set to the number of image tiles, i.e. $K=16$, we find that some of the clustered subdomains are quite similar to each other. The repetition of discovered subdomains and the degraded performance suggest that similar style patterns are shared among different image tiles and enforcing image tile-wise style separation is inappropriate.
When implementing our method, we set $K=10$ for all experiments.

Additionally, we have evaluated the performance for different types of clustering methods\,(i.e., Mean-Shift algorithm\;\citep{cheng1995mean}) by replacing the K-Means algorithm adopted in Section\;\ref{sec:kmeans}. 
On the BBBC039$\rightarrow$Kumar adaptation benchmark, the Mean-Shift algorithm attains inferior results in terms of all evaluation metrics\;(PQ=0.5194, AJI=0.5523, DICE=0.7917) compared with K-Means.
We observe that the performance degradation is as a consequence of the biased cluster split estimation.
For the Mean-Shift algorithm, it leverages region searching to automatically determine the number of clusters.
However, in the domain of histopathology, the data distribution intrinsically lies on a high-dimensional manifold with intricate levels of representation hierarchies.
Density-based clustering algorithms like Mean-Shift inevitably suffer from erratic subspace organization and structuring, deteriorating the representativeness of identified subpartitions.

\subsection{Effectiveness of the nucleus shape and structure preserving module}
\label{sec:effect_of_shape_preserve}
\begin{figure}[!t]
\centerline{\includegraphics[width=0.6\columnwidth]{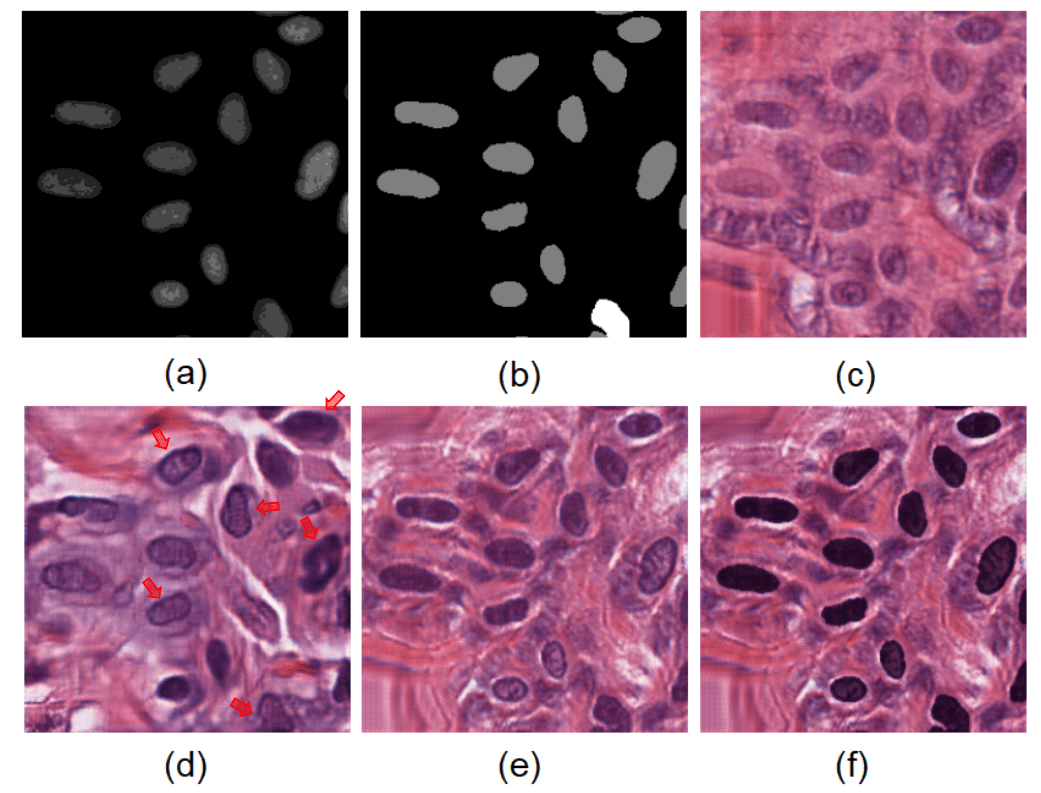}}
\caption{Visual comparison for image translation results of different methods from fluorescence microscopy to histopathology images. (a)\;source fluorescence microscopy image patch, (b)\;corresponding nuclei annotations, image synthesized using (c)\;CycleGAN, (d)\;our method w/o shape regularization, (e)\;our method with shape regularization,  (f)\;plot segmentation masks on top of (e).
	In (b), different colors are used to distinguish touching nuclei.}
\label{fig:shape_preserve}
\end{figure}
\begin{figure}[!t]
\centerline{\includegraphics[width=0.7\columnwidth]{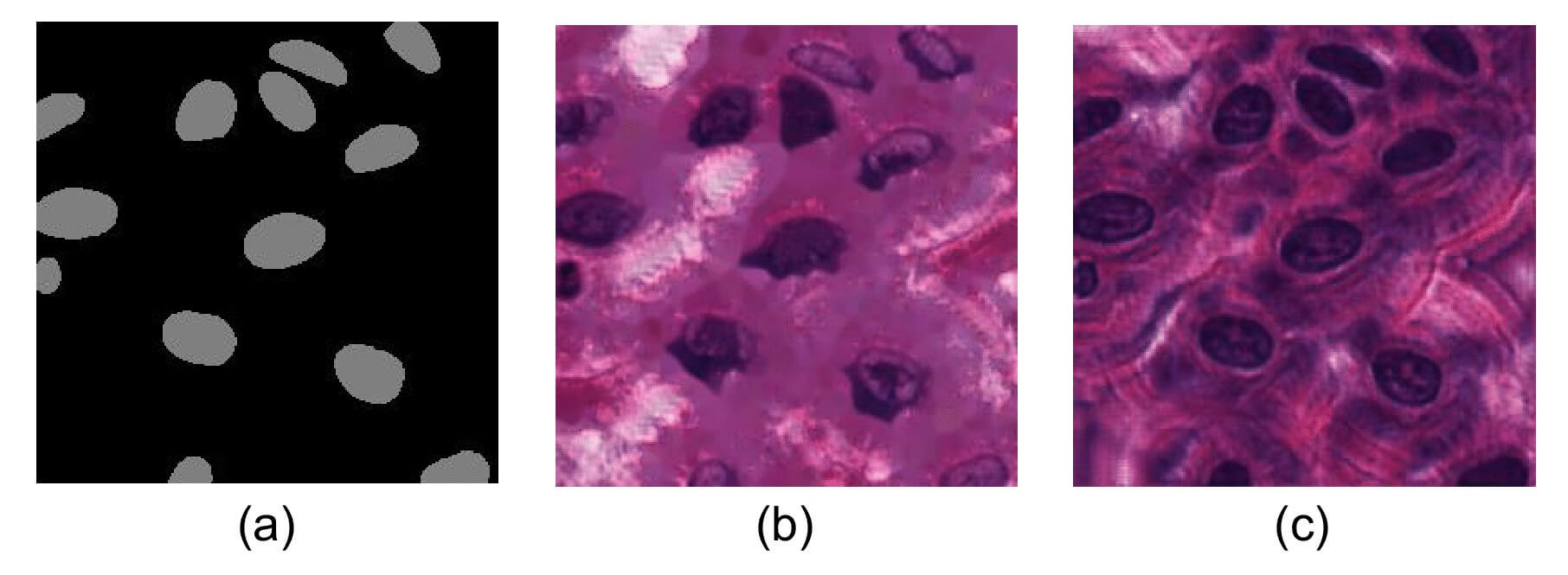}}
\caption{Illustration of the detrimental effects of nuclei inpainting and our improvements. (a)\;corresponding nuclei annotations, image synthesized using (b)\;CycleGAN along with nuclei inpainting, (c)\;our method.}
\label{fig:against_inpaint}
\end{figure}
In cross-domain nuclei instance segmentation, due to the lack of nucleus object-level supervision, conventional image translation models such as CycleGAN and DRIT suffer from two weaknesses, namely nucleus over-generation and deformation. Firstly, as depicted in Fig.\;\ref{fig:shape_preserve}(c), the histopathology image synthesized with CycleGAN contains inappropriate nuclei objects which do not exist in the raw fluorescence microscopy image. In the previous works\;\citep{Liu2020pdam}, an auxiliary object inpainting mechanism is introduced to tackle the issue. However, in this way, substantial background textures with rich semantic information are wiped out, as shown in Fig.\;\ref{fig:against_inpaint}(b).
In addition, as noted in Fig.\;\ref{fig:shape_preserve}(d), the synthesized image fails to preserve the nucleus shape and structure details and results in boundary deformation.

Dedicated to overcoming those obstacles, we design the nucleus shape and structure preserving module to enforce consistency on both semantic masks and object boundaries in image translation. Fig.\;\ref{fig:against_inpaint}(c) shows that our method is able to synthesize precisely matched histopathology images. As no post-calibration is required, background texture pattern can be left uncontaminated. Furthermore, from Fig.\;\ref{fig:shape_preserve}(d)-(f), it can be seen that owing to the proposed module, our method almost perfectly preserves the nucleus shape details. The performance drop when removing the module\;($2^{nd}$ row, Table\;\ref{tab:ablation_study}) also confirms its importance.

To jointly measure the contributions of modules in Stage I, we conduct a comparative analysis by replacing our proposed image translation framework with existing disentanglement-based approach, i.e., DRIT\;\citep{Lee2020drit} and DRANet\;\citep{lee2021dranet}. The experimental results are presented in the $6^{th}$ and $7^{th}$ rows in Table\;\ref{tab:ablation_study}.
It is observed that our framework outperforms all the competitive approaches by a large margin in terms of all evaluation metrics, which in turn verifies the efficacy to progressively model the inner structure of the heterogeneous histopathology domain and enforce semantic preservation of nucleus structures.

\subsection{Effectiveness of instance-level disentanglement and style consistency}
In Stage I, the disentanglement is conducted at the global level to characterize image visual pattern variations and thus enable controllable image style transfer and synthesize diverse target-like patches. While in Stage II, the disentanglement is accomplished in the local instance-level and the purpose is to acquire domain-invariant feature representations.
As shown in $3^{rd}-4^{th}$ row, Table\;\ref{tab:ablation_study}, compared with the global-level disentanglement-only lower bound, our method with instance-level disentanglement as well as ROI reconstruction consistency constraint promotes the average PQ to 0.5287. 
Then by additionally introducing the global-local style consistency module to facilitate the unbiased modeling of subdomain-specific characteristics, our method is able to achieve an average PQ of 0.5527, which improves the instance segmentation performance by $4.4\%$ compared with W/o instance-level disentanglement.
The remarkable increase in instance segmentation accuracy confirms that explicitly formulating precise and distinctive domain-specific instance-level attributes is beneficial for the separation of domain-invariant representations from the highly-entangled feature maps and, as a result, fosters cross-domain adaptation.
In comparison, as indicated in $5^{th}$ row, Table\;\ref{tab:ablation_study}, when the clustering-based pseudo subdomain label strategy employed in Stage I is similarly introduced in Stage II, a significant performance drop is observed. It backs up our claim that on account of the category-wise nuclei heterogeneity, simply encouraging instance appearance attributes to form subdomain-wise clusters would result in biased feature disentanglement and is detrimental to domain adaptation.

\begin{figure}[!t]
	\centerline{\includegraphics[width=\columnwidth]{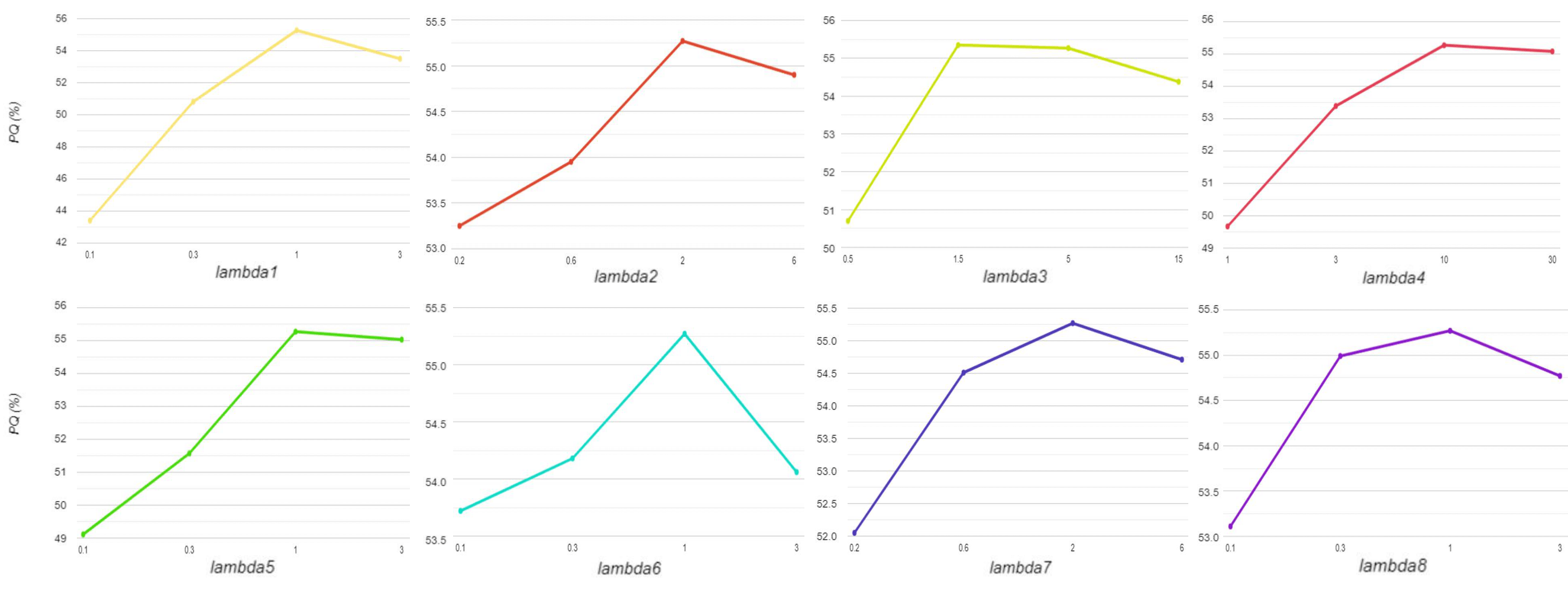}}
	\caption{Sensitivity Analysis of Loss Weighting Terms $\lambda_{1\sim8}$ on the BBBC039 to Kumar Benchmark.}
	\label{fig:loss_sensitivity}
\end{figure}
\subsection{Sensitivity analysis on loss weighting terms}
To further investigate how the choice of loss weighting terms impact the overall performance of the proposed method, we perform sensitivity analysis on those terms.
	Specifically, for each weighting term defined in Equations 4 and 6, we firstly set it according to the parameter configuration presented in Section\;4.2.
	Then, we scale it by a factor $0.1, 0.3$, and $3$, respectively.
	The corresponding quantitative UDA nuclei instance segmentation performance on the BBBC039 to Kumar benchmark is presented in Fig.\;\ref{fig:loss_sensitivity}.
	It is observed that weighing terms corresponding to fundamental purposes\;(e.g., DRIT adversarial and image reconstruction losses in Stage I - $\lambda_1$, and Mask R-CNN instance segmentation losses in Stage II - $\lambda_5$) could have a more significant impact on the final results.
	In addition, the figure indicates that the overall performance is not sensitive to the specific choices of those weighing terms as long as they are set within a reasonable interval\;(i.e., scale factor lies in $0.3\sim3$).

\begin{table}[!t]
	\centering
	\fontsize{7}{8}\selectfont
	\begin{threeparttable}
		\caption{Evaluations for cross-stain nuclei classification and instance segmentation on the CoNSep$\rightarrow$PanNuke adaptation benchmark.}
		\label{tab:cross_stain_extend1}
		\setlength{\tabcolsep}{1.6mm}{
			\begin{tabular}{ccccccccc}
				\toprule
				\multirow{2}{*}{\textbf{Method}}&
				\multicolumn{4}{c}{\textbf{Classification}}&
				\multicolumn{4}{c}{\textbf{Instance Segmentation}}\cr
				\cmidrule(lr){2-5}\cmidrule(lr){6-9}
				&Epi.&Inf.&Con.&\textbf{Avg.}&Epi.&Inf.&Con.&\textbf{Avg.}\cr
				\midrule				
				DARCNN\;\citep{Hsu2021}
				&$0.6584$&$0.5719$&$0.5930$&$0.6078$
				&$0.2611$&$0.2530$&$0.1917$&$0.2353$\cr
				PDAM\;\citep{Liu2020pdam}
				&$0.6840$&$0.5804$&$0.5878$&$0.6174$
				&$0.2804$&$0.2930$&$0.1937$&$0.2557$\cr
				DHA\;\citep{Park2020}
				&$0.6526$&$0.5669$&$0.5794$&$0.5996$
				&$0.2548$&$0.2681$&$0.1812$&$0.2347$\cr
				Ours
				&$\bf{0.7193}$&$\bf{0.6103}$&$\bf{0.6142}$&$\bf{0.6479}$
				&$\bf{0.2915}$&$\bf{0.3213}$&$\bf{0.1991}$&$\bf{0.2706}$\cr
				\bottomrule
		\end{tabular}}
	\end{threeparttable}
\end{table}
\begin{table}[!t]
\centering
\fontsize{7}{8}\selectfont
\begin{threeparttable}
	\caption{Evaluations for cross-stain nuclei classification and instance segmentation on the GlaS$\rightarrow$Dpath adaptation benchmark.}
	\label{tab:cross_stain_extend2}
	\setlength{\tabcolsep}{1.6mm}{
		\begin{tabular}{ccccccccc}
			\toprule
			\multirow{2}{*}{\textbf{Method}}&
			\multicolumn{4}{c}{\textbf{Classification}}&
			\multicolumn{4}{c}{\textbf{Instance Segmentation}}\cr
			\cmidrule(lr){2-5}\cmidrule(lr){6-9}
			&Epi.&Inf.&Con.&\textbf{Avg.}&Epi.&Inf.&Con.&\textbf{Avg.}\cr
			\midrule				
			DARCNN\;\citep{Hsu2021}
			&$0.7126$&$0.6227$&$0.4795$&$0.6049$
			&$0.2809$&$0.3676$&$0.2310$&$0.2932$\cr
			PDAM\;\citep{Liu2020pdam}
			&$0.7739$&$0.6764$&$0.5045$&$0.6516$
			&$0.3152$&$0.4156$&$0.2531$&$0.3280$\cr
			DHA\;\citep{Park2020}
			&$0.7206$&$0.6614$&$0.5092$&$0.6304$
			&$0.2880$&$0.4078$&$0.2592$&$0.3183$\cr
			Ours
			&$\bf{0.7945}$&$\bf{0.7022}$&$\bf{0.5379}$&$\bf{0.6782}$
			&$\bf{0.3374}$&$\bf{0.4298}$&$\bf{0.2801}$&$\bf{0.3491}$\cr
			\bottomrule
	\end{tabular}}
\end{threeparttable}
\end{table}
\subsection{Evaluations on class-aware cross-stain evaluation}
To verify the effectiveness and robustness of our method under diverse domain-adaptive scenarios and problematic formulation, we perform extended evaluations on two cross-stain settings with the aim to not only delineate the boundary of each nucleus but also identify its functional type.
Specifically, we firstly consider the adaptation from CoNSep\;\citep{Graham2019hover} to PanNuke\;\citep{gamper2019pannuke}. 
Those two benchmarks are constructed with histopathology imaging data collected different countries and institutes, with domain shifts inherently present due to inconsistency in staining procedure.
PanNuke dataset is composed of histopathology tiles sampled from a broad range of organs and cancers, conforming to the heterogeneity proposition in target domain.
Next, we perform evaluations on the adaptation setting from GlaS to Dpath\;\citep{graham2021lizard}, for which the two datasets are collected from cohorts across different countries with evident data distribution shifts.
The quantitative comparative results are presented in Table\;\ref{tab:cross_stain_extend1}, where we adopt the F1 metric to demonstrate classification accuracy and use the class-wise PQ score to indicate segmentation performance for each type of nucleus.
Epi., Inf., and Con. correspond to the nuclei of epithelial, inflammatory, and connective cells, respectively.
Avg. denotes the class-averaged results.
It is demonstrated that our proposed method consistently outperforms the competing methods in terms of all evaluation metrics over two technical tasks.
The experimental comparison results justify the effectiveness and general applicability of our method in addressing miscellaneous data distribution shifts in the histopathology domain under different levels of output demands.

\subsection{Visualization results of style randomization}
\begin{figure}[!t]
\centerline{\includegraphics[width=0.9\columnwidth]{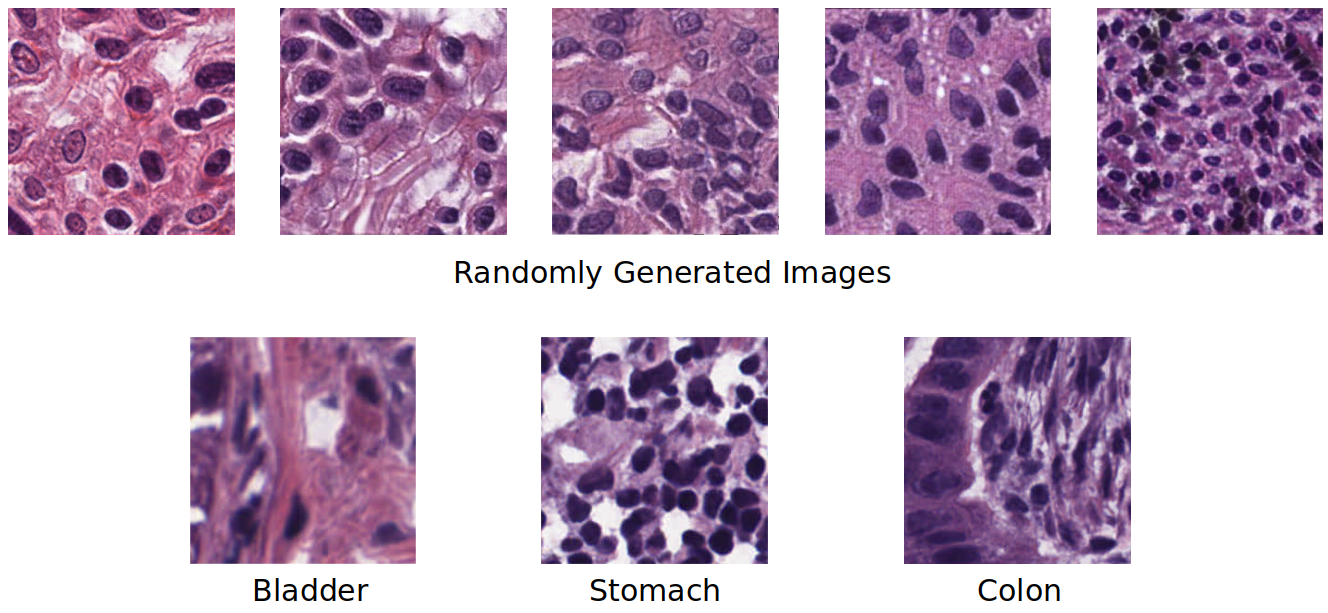}}
\caption{Visualization results of style randomization.
	Images in the first row are examples of the images generated with randomly sampled style attribute vectors.
	Images in the second row are the real image patches from unseen testing subdomains\;(i.e., Bladder, Stomach, and Colon).
	All images are from the BBBC039 to Kumar benchmark.}
\label{fig:randomization}
\end{figure}

In Fig.\;\ref{fig:randomization}, we present several examples of the images generated with randomly sampled style attribute vectors.
It can be observed that compared with images shown in Fig.\;\ref{fig:visual_subdomains}, which are sampled from the Kumar training set\;(seen subdomains), those randomly synthesized images demonstrate more significant visual similarity to the unseen testing ones.
This finding explains why our method could remarkably surpass the competing ones regarding the generalization capability to unseen subdomains, which is showcased in Table\;\ref{tab:seen_unseen_comparison}.

In addition, it is noted that generating images similar to unseen subdomains is not necessary to strengthen the model's generalization capability.
\cite{Yamashita2021learning} argued that style transfer with arbitrary style sources, including the ones divergent from the task domain, could enhance the model's robustness against domain shifts.
They utilized artistic paintings as style sources and performed style transfer to augment histopathology images.
The synthesized images are apparently dissimilar to the real histopathology images, yet they still attain substantial improvements when testing the trained model on unseen subsets of histopathology images.
It indicates that the core to style augmentation is the learning of domain-invariant visual representations, instead of generating images similar to unseen subdomains.

\subsection{Impacts of color normalization}
Color normalization is a common approach to alleviate the heterogeneity of histopathology images, which is incurred by inconsistent acquisition, processing, and staining procedures \cite{vsvihlik2015color, farahani2022deep}.
In this section, we employ two color normalization techniques to the Kumar dataset and then evaluate how the proposed method performs on those normalized images.

Specifically, we introduce two widely used color normalization methods, namely RGB histogram specification\;\citep{Gonzalez2002} and color transfer\;\citep{Reinhard2001color}.
We randomly select one image as the reference image and accordingly normalize all the images in the Kumar dataset.
Experiments are then conducted on the BBBC039 to Kumar benchmark.
However, it is observed that those color normalization methods have negative effects on the results.

\begin{figure}[!t]
\centerline{\includegraphics[width=\columnwidth]{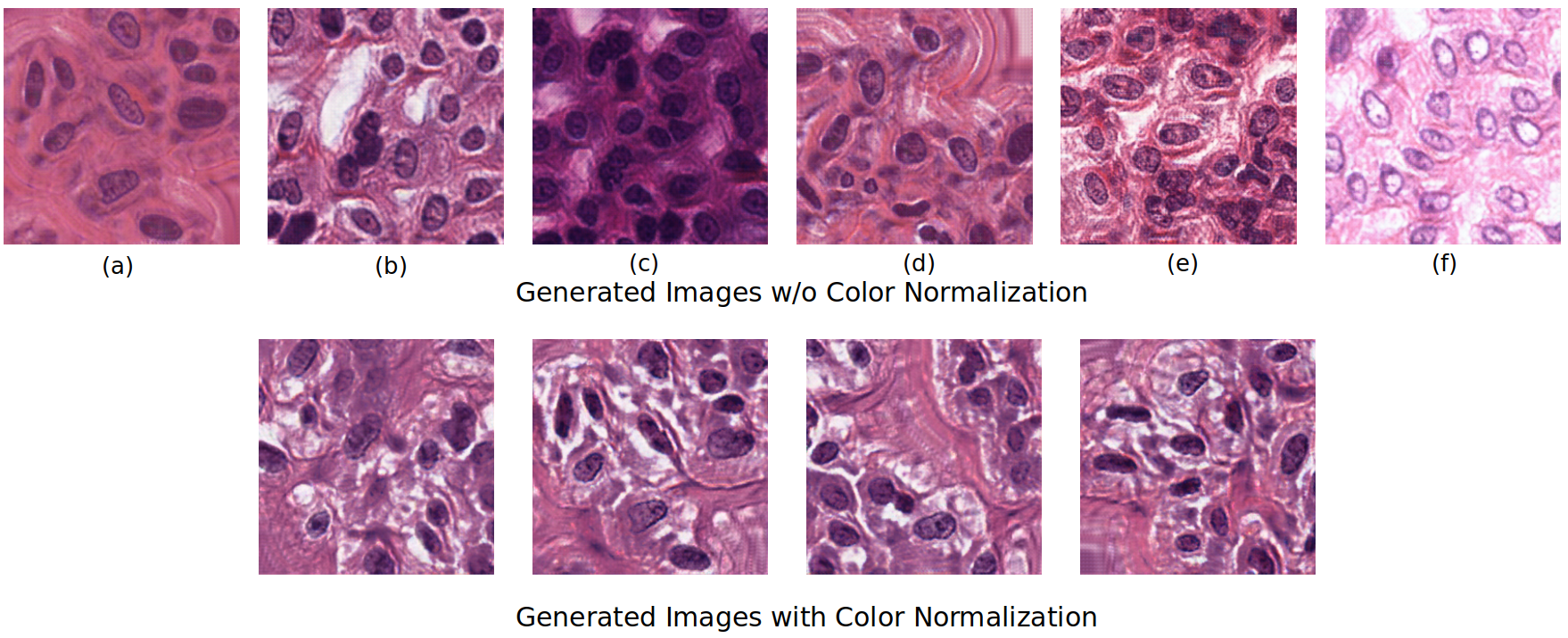}}
\caption{Examples of images generated in Stage I with and w/o color normalization\;(RGB histogram specification).}
\label{fig:stage1_failure}
\end{figure}
First, we find that in Stage I, the diversity of generated images is limited, as shown in Fig. \ref{fig:stage1_failure}.
Here, the image diversity not only denotes the global color distribution, but also indicates other informative cancer-specific characteristics, such as the texture of nuclei and biological tissues.
For instance, the nuclei in Fig. \ref{fig:stage1_failure}(e)(f) possess distinct texture patterns compared with (a)-(d), whereas all nuclei generated with color normalization are very similar to each other.
It reveals that the cancer-specific attributes are closely entangled with color patterns.
Color normalization is indeed helpful to reduce the color discrepancy caused by stain variation, but it would also inevitably erase the cancer-specific attributes and is detrimental to the diversity of translated images, which consequently leads to poor cross-domain performance.
And no matter how we choose the value of K, the performance is always inferior. It is because the cancer-specific attributes are removed, and the extracted style features cannot represent the unique characteristics of different cancer types. In this case, the clustering of subdomains in Section 3.1.2 is meaningless.

\begin{figure}[!t]
\centerline{\includegraphics[width=\columnwidth]{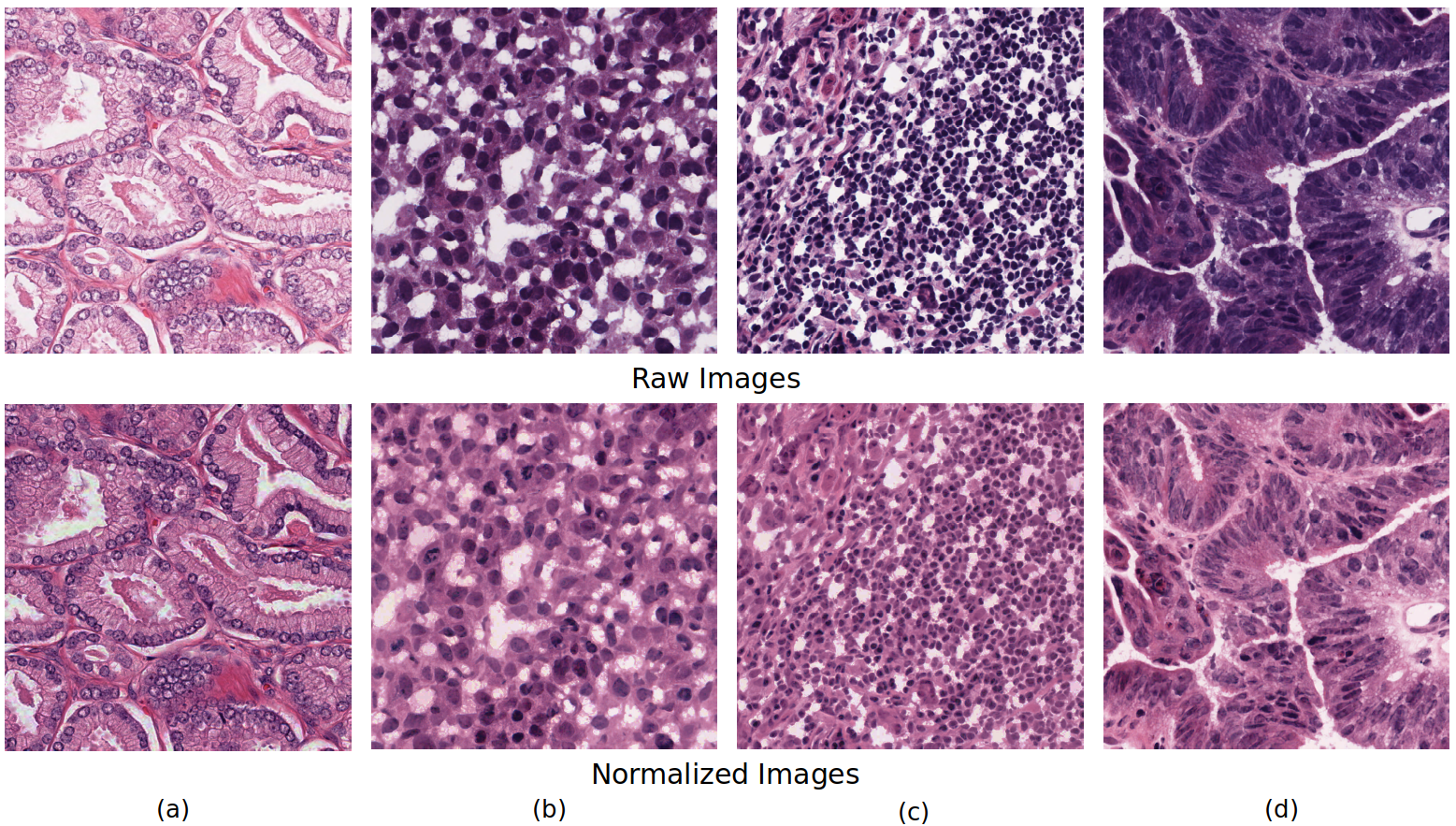}}
\caption{Examples of failure cases with color normalization.}
\label{fig:norm_failure}
\end{figure}

Second, we find that after Stage II, color normalization in fact deteriorates the overall accuracy of cross-domain nuclei instance segmentation.
The quantitative results are presented in Table\;\ref{tab:color_norm}.
\begin{table}[!t]
\label{tab:norm_overall}
\centering
\fontsize{7}{8}\selectfont
\begin{threeparttable}
	\caption{Quantitative analysis on the impacts of color normalization on the BBBC039 to Kumar benchmark.}
	\label{tab:color_norm}
	\setlength{\tabcolsep}{3.5mm}{
		\begin{tabular}{ccc}
			\toprule
			%\multirow{2}{*}{Normalization Method}&
			%\multicolumn{2}{c}{PQ}\cr
			%\cmidrule(lr){2-3}
			Normalization Method&PQ&AJI\cr
			\midrule		
			RGB histogram specification&$0.4402$&$0.4978$\cr		
			Color transfer&$0.4665$&$0.5066$\cr
			w/o normalization&$0.5527$&$0.5797$\cr
			\bottomrule
	\end{tabular}}
\end{threeparttable}
\end{table}
We also notice that the performance drop is especially severe for image tiles shown in Fig. \ref{fig:norm_failure}.
The images in the first column (a) are from prostate\;(seen subdomain), and the images in the other three columns (b)-(d) are from bladder, stomach, and colon\;(unseen subdomains).
This observation substantiates our statement.
When color normalization is not employed, our proposed method can synthesize images with various texture patterns, including the unique nuclei texture in prostate images Fig. \ref{fig:norm_failure}(a)\;(the corresponding synthesized image is Fig. \ref{fig:stage1_failure}(f)).
The diversity of generated images also contributes to its success on unseen subdomains, as discussed in Section 5.5.
On the contrary, with color normalization, the synthesized nuclei are of unitary texture patterns.
It compromises the Stage II model's generalization capability and results in inferior overall performance.

\section{Conclusion}
Data distribution heterogeneity across various cancer types and sampling tissues arises as the major obstacle undermining the potential of applying UDA methods to facilitate digital pathology.
In this paper, we present the first work to explicitly consider the composited nature of data distribution in the histopathology domain and thereby develop a holistic framework to rectify the biased alignment procedure along adaptation.
With the aim of inducing well-regulated decomposition between the informative pathological attributes and the confounding modality/stain-specific factors, we propose the progressive subdomain partition and cross-scale co-regularization strategies.
Those key components collaboratively shape the embedding space where domain-invariant structural content can be decoupled from the task-irrelevant distributional variance.
For empirical evaluations of our method, we perform extensive experiments over a diverse set of cross-modality and cross-stain adaptation benchmarks to verify its effectiveness and broader applicability.
The quantitative and qualitative comparison results demonstrate the superiority of our method over state-of-the-art UDA and OCDA approaches in various evaluation metrics across different tasks.

\section*{Data Availability Statement}

The datasets generated during and/or analysed during the current study are available in the following repositories:
\begin{itemize} 
\item https://bbbc.broadinstitute.org/BBBC039 
\item https://drive.google.com/drive/folders/1l55cv3DuY-f7-JotDN7N5nbNnjbLWchK
\item https://www.dropbox.com/sh/e7oz4nhp3gekvk4/AAC-xuqg5DUx0H5JdqPApbWTa?dl=0s
\end{itemize}


\bibliography{sn-bibliography-final}% common bib file


\end{document}